%% file: usenix_main.tex
\documentclass[letterpaper,twocolumn,10pt]{article}
\usepackage{usenix-2020-09}
\usepackage{graphicx} % allows inclusion of PDF, PNG and JPG images
\usepackage{todonotes}
\usepackage{algorithm} 
\usepackage{algpseudocode} 
\usepackage{listings}
\usepackage{xcolor}
\usepackage{multirow}
\usepackage{adjustbox}
\usepackage{float}
\usepackage{tcolorbox}
\usepackage{subcaption}
\usepackage{verbatim}
\usepackage{tcolorbox}
\usepackage{subcaption}
\usepackage{amsmath}
\usepackage{amsthm}
\usepackage{amssymb}
\usepackage{amsfonts}
\usepackage{subfloat}

\newcommand{\eg}{\textit{e.g\@.}}

\newcommand{\ie}{\textit{i.e\@.}}

\newcommand{\myexample}[2]{
    \begin{tcolorbox}[colback=black!5!white,colframe=black,title={#1}]
        #2
    \end{tcolorbox}
}

\definecolor{codegreen}{rgb}{0,0.6,0}
\definecolor{codegray}{rgb}{0.5,0.5,0.5}
\definecolor{codepurple}{rgb}{0.58,0,0.82}
\definecolor{backcolour}{rgb}{0.95,0.95,0.92}

\lstdefinestyle{mystyle}{
	backgroundcolor=\color{backcolour},   
	commentstyle=\color{codegreen},
	keywordstyle=\color{magenta},
	numberstyle=\tiny\color{codegray},
	stringstyle=\color{codepurple},
	basicstyle=\ttfamily\footnotesize,
	breakatwhitespace=false,         
	breaklines=true,                 
	captionpos=b,                    
	keepspaces=true,                 
	numbers=left,                    
	numbersep=5pt,                  
	showspaces=false,                
	showstringspaces=false,
	showtabs=false,                  
	tabsize=2
}

\usepackage{tikz}
\usepackage{amsmath}
\usepackage{booktabs} % To thicken table lines
\usepackage{cleveref}
\newcommand{\RS}{\textit{Randomised Smoothing}}

\begin{document}
%-------------------------------------------------------------------------------

%don't want date printed
\date{}
% Randomness-Based Attacks and Vulnerabilities in Machine Learning: A Study on Randomised Smoothing
\title{\Large \bf Machine Learning needs Better Randomness Standards:\\~\RS~and \texttt{PRNG}-based attacks}

%for single author (just remove % characters)
\author{
% Anonymous authors\\
% Anonymous universities
{\rm Pranav Dahiya}\\
University of Cambridge
\and
{\rm Ilia Shumailov}\\
University of Oxford
\and
{\rm Ross Anderson}\\
University of Cambridge\\ \& University of Edinburgh
}

\maketitle

\input{sections/abstract}

\input{sections/intro}

\input{sections/related}
\input{sections/methodology}
\input{sections/evaluation}
\input{sections/discussion}

\input{sections/conclusion}

%-------------------------------------------------------------------------------

% Fix references in bibliography

\bibliographystyle{abbrv}
\bibliography{bibliography}

\input{sections/appendix}

%%%%%%%%%%%%%%%%%%%%%%%%%%%%%%%%%%%%%%%%%%%%%%%%%%%%%%%%%%%%%%%%%%%%%%%%%%%%%%%%
\end{document}

%% file: sections/abstract.tex
\begin{abstract}

Randomness supports many critical functions in the field of machine learning (ML) including optimisation, data selection, privacy, and security. ML systems outsource the task of generating or harvesting randomness to the compiler, the cloud service provider or elsewhere in the toolchain. Yet there is a long history of attackers exploiting poor randomness, or even creating it -- as when the NSA put backdoors in random number generators to break cryptography. In this paper we consider whether attackers can compromise an ML system using only the randomness on which they commonly rely. We focus our effort on~\RS, a popular approach to train certifiably robust models, and to certify specific input datapoints of an arbitrary model. We choose~\RS~since it is used for both security and safety -- to counteract adversarial examples and quantify uncertainty respectively. Under the hood, it relies on sampling Gaussian noise to explore the volume around a data point to certify that a model is not vulnerable to adversarial examples. We demonstrate an entirely novel attack, where an attacker backdoors the supplied randomness to falsely certify either an overestimate or an underestimate of robustness for up to 81 times. We demonstrate that such attacks are possible, that they require very small changes to randomness to succeed, and that they are hard to detect. As an example, we hide an attack in the random number generator and show that the randomness tests suggested by NIST fail to detect it. We advocate updating the NIST guidelines on random number testing to make them more appropriate for safety-critical and security-critical machine-learning applications. 
\end{abstract}

%% file: sections/intro.tex
\section{Introduction}

Randomness is crucial in machine learning (ML), serving a number of  purposes across different areas. 
One use case is in federated learning, where it helps user selection to ensure a diverse representation of participants and prevent bias~\cite{konecny2017federated}; it can also provide privacy amplification in the process~\cite{balle2018privacy}.
It is heavily used in optimisation, providing the foundation for Stochastic Gradient Descent~\cite{robbins1951stochastic}, as well as generally for data sampling, enabling the selection of representative subsets for model training~\cite{mehrabi2022survey}. It enables Monte Carlo methods~\cite{metropolis1949monte}. It is essential in active learning, a process where an algorithm actively selects the most informative samples for labelling, in order to reduce the training cost~\cite{kirsch2019batchbald}. It forms the basis of differential privacy, the de facto standard for quantifying privacy, where noise is added to data in such a way as to protect individual privacy while still allowing for accurate analysis~\cite{dwork2014algorithmic}. It contributes to the generation of synthetic data, which aids in expanding the training set and enhancing the generalisation capabilities of ML models~\cite{kurakin2023harnessing}. In short, randomness is widely used and highly significant for ML.

Yet little thought  has been given to the vulnerabilities that might result from weak randomness. We therefore explore the extent to which an attacker can change the safety-critical decision-making of a target system, purely by exploiting or tinkering with random number generators. We focus our efforts on \RS, a standard technique for quantifying uncertainty\footnote{The literature does not view~\RS~as an uncertainty estimation technique, yet it is one when it is used to certify a prediction around a target datapoint.} in a given blackbox model prediction~\cite{cohen_certified_2019}. This is heavily used in practice to combat adversarial examples; it can even be used to provide robustness certification. It samples isotropic Gaussian noise and adds it to a critical datapoint in order to measure the probability that the addition of  noise causes the model prediction to change, leading to a spurious decision . Sampling high-quality noise is crucial for this purpose, yet in practice we rarely check how normal our Gaussian noise is.

In this paper, we construct two attacks against~\RS~that assume an attacker can influence the random number generator on which the model relies. The first is a naive attack that simply replaces a Gaussian distribution with a different distribution~\eg~Laplace noise. This disrupts confidence quantification significantly for both over- or under-estimation, but is relatively easy to detect. We present the attack intuition visually in~\Cref{fig:example_figure}. Following this naive proof of concept, we construct a more powerful and covert attack: a bit-flipping \texttt{PRNG} attacker that changes only a single bit out of every 64 bits deep in the implementation of the random number generator. We show that this can cause mis-quantification of the true confidence by up to a factor of $\times 81$ and is extremely hard to detect. This change to the \texttt{PRNG} is covert, in that it does not cause it to fail the official NIST suite of randomness tests. 

This highlights the inadequacy of current standards and defences in protecting against attacks targeting randomness in machine learning. We argue that similar randomness-based attacks can be devised against other ML techniques, such as differential privacy. It follows that the standards are insufficient to guarantee the security and privacy of machine learning systems in the face of sophisticated adversaries. We then discuss practical ways of tackling these vulnerabilities. We aim to empower researchers and practitioners to understand the extent to which they place their trust in their toolchain's source of randomness, and develop more robust defences against the abuse of this trust.

By developing an attack that exploits randomness and demonstrating its impact on the mechanisms most widely used to certify safety properties in ML, we expose the need for improved standards. By exploring potential mitigations, we hope to enable people to build more secure and resilient machine-learning systems. \\

In this paper, we make the following contributions:

\begin{itemize}
    \item We demonstrate a new class of attacks against \RS~based on the substitution of an underlying noise distribution.
    \item We show how an attacker can change the underlying randomness generators to defeat it more covertly and even more effectively.
    \item We show that NIST's randomness tests with default parameters fail to catch our attacks and argue for updated randomness standards.
\end{itemize}

%% file: sections/related.tex
\section{Related work}

\begin{figure}[t]
    \centering
    \includegraphics[width=\linewidth]{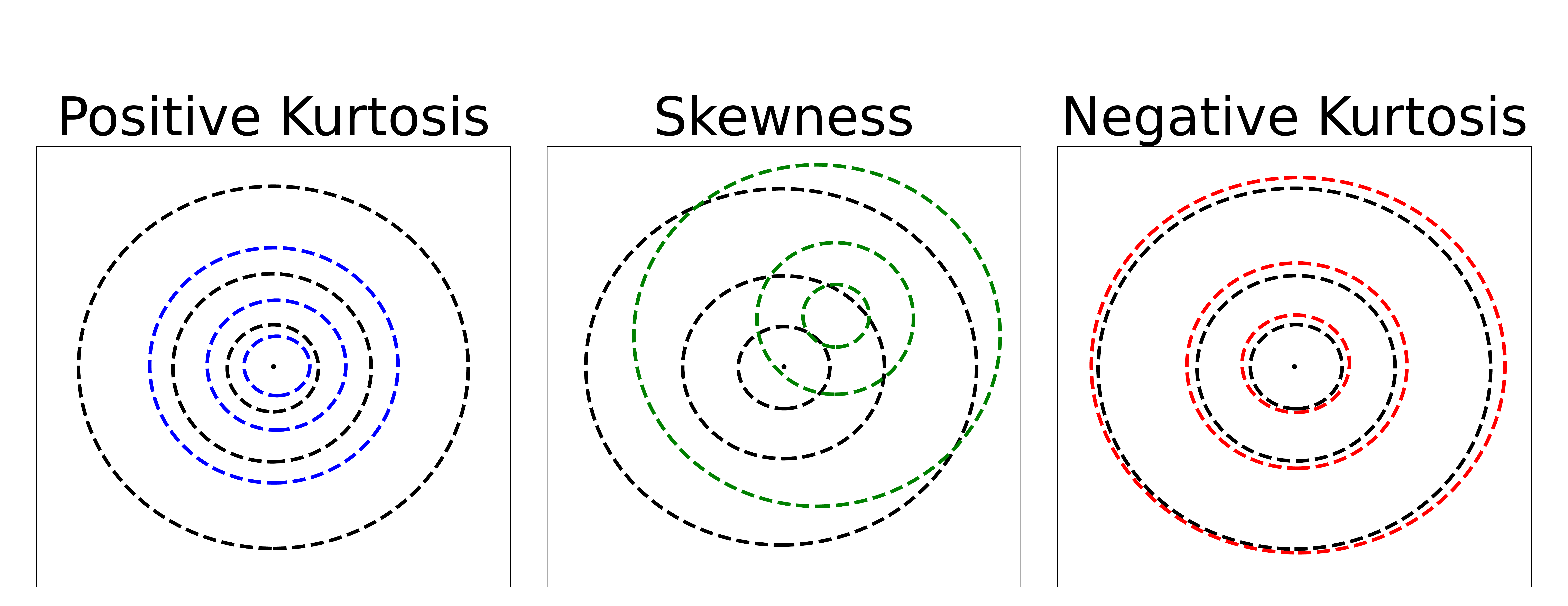}
    \caption{A visual example of how manipulated randomness affects~\RS. Here we aim to certify a point in the middle. To do that we sample data around it using noise -- a normal noise sampling for~\RS~is shown in black. Manipulated random noise with lower kurtosis, skewness and higher kurtosis are shown in order from left to right. Manipulated noise misrepresents the true decision space and leads to incorrect prediction confidence.}
    \label{fig:example_figure}
\end{figure}

In this section we cover the work that relates to the class of attacks we developed. \Cref{sec:related:advx} covers adversarial examples and~\Cref{sec:related:randomisedsmooth} discusses how~\RS~can use randomness to counteract them for both safety and security. It is followed by a section on randomness. 

\subsection{Adversarial Examples}
\label{sec:related:advx}
The evasion attack problem was formulated as an optimisation problem by Biggio et al.~\cite{biggio_evasion_2013} in 2013. Szegedy et al.~\cite{szegedy_intriguing_2013} demonstrated that deep learning models are indeed highly susceptible to evasion attacks. To determine the optimal adversarial perturbation on a given input $x^0$, an attacker aims to minimise $L :\ \mathcal{X} \rightarrow \mathbb{R}$, where $L$ is the model's discriminant function~\cite{biggio_evasion_2013}:

% \begin{equation}
% 	\label{eqn:evasion-attack}
% 	\textbf{x}^* = \underset{x}{\operatorname{arg min}}\ f(\textbf{x})\ \text{where}\ d(\textbf{x}, \textbf{x}^0) \leq d_{max}.
% \end{equation}

\begin{equation}
\label{eqn:evasion-attack}
x^* = \underset{x:\;d\left(x,x^0\right) \leq d_{max}} {\operatorname{argmin}}L(x).
\end{equation}

The choice of the distance function $d: \mathcal{X} \times \mathcal{X} \rightarrow \mathbb{R}$ is domain-specific, and the maximum size of the perturbation is limited to $d_{max}$. More efficient attacks now exist~\cite{carlini_adversarial_2017,croce2020reliable}.

\subsection{Certification in Machine Learning}
\label{sec:related:certification}

Certification techniques attempt to invert \Cref{eqn:evasion-attack} to determine the maximum perturbation radius around an input, inside which no adversarial examples exist. This is different from what the security community means by certification in the context of formal verification \cite{cullen_exploiting_2023}: that refers to programmatic proof that a piece of software performs exactly as intended, matching some high-level abstraction. For example, a certified compiler will always correctly translate legal code correctly into its compiled binary or throw an appropriate error \cite{leroy_formal_2009}. If the high-level abstraction of a machine learning model is considered to be the function $f: \mathcal{X} \rightarrow \mathcal{Y}$, mapping inputs to their true classes, then certification here would imply that, for a given volume around a point from $\mathcal{X}$, the prediction does not change from $\mathcal{Y}$. Many certification techniques in current use are approximate and, despite the fact that theorems may be offered around their behaviour, they only provide probabilistic guarantees for the absence of adversarial examples in an $\epsilon$ ball around a given input~\cite{croce2020reliable}. 

\RS~is not the only certification technique. Other mechanisms exist, \eg~by constructing a bounding polytope instead of a hypersphere, or using either linear relaxation or interval bound propagation~\cite{xu_automatically_2016}. However, these require changes to the training process and model re-engineering, which in turn makes them hard to scale up to more complex model architectures and higher dimensional data~\cite{cullen_exploiting_2023}. \RS~is one of the most widely used certification techniques and it underpins current state-of-the-art certification algorithm for machine learning models~\cite{zhang2023diffsmooth}. It can be applied to any underlying model in a black-box manner and scaled up to larger model architectures. It is used both for security, \ie~to combat adversarial examples, and safety, \ie~to provide confidence in predictions. That is why we made it the target of this work.
\subsection{Randomised Smoothing}
\label{sec:related:randomisedsmooth}

Any classifier $f$ can be used to construct a smoothed classifier $g$ such that:
\begin{equation}
	\label{eqn:smoothed-f}
	g(x) = \underset{c \in \mathcal{Y}}{\operatorname{arg\ max}}\ \mathbb{P}\left\{ f(x + \epsilon) = c \right\},\ 
	\text{where}\ \epsilon \sim \mathcal{N}(0, \sigma^2\mathit{I}).
\end{equation}

This formulation allows for easy approximation of $g$ for a given confidence bound using a Monte Carlo algorithm by sampling $\epsilon$ from $\mathcal{N}(0, \sigma^2 \mathit{I})$ \cite{cohen_certified_2019}. While previous work used differential privacy \cite{dwork_calibrating_2006,lecuyer_certified_2019} and Renyi divergence \cite{van_erven_renyi_2014,li_second-order_2018} to determine the radius of the hypersphere around an input inside which the absence of adversarial examples can be provably verified, Cohen et al.~\cite{cohen_certified_2019} used the Neyman-Pearson lemma \cite{neyman_ix_1933} to obtain a certification radius which is also provably maximal. The certified radius obtained from randomised smoothing is:

\begin{equation}
	\label{eqn:radius}
	R = \frac{\sigma}{2} \left( \Phi^{-1}(p_A) - \Phi^{-1}(p_B) \right),
\end{equation}

where $\Phi^{-1}$ is inverse of the standard normal cumulative distribution function, $p_A$ is the probability of the most probable class $c_A \in \mathcal{Y}$ and $p_B$ is the probability of the next most probable class. This result holds true even if $p_A$ and $p_B$ are replaced with $\underline{p_A}$, a lower bound on $p_A$, and $\overline{p_B}$, an upper bound on $p_B$ such that $p_A \geq \underline{p_A} \geq \overline{p_B} \geq p_B$. The mathematical proof for~\Cref{eqn:radius} and its maximality can be found in the appendices of the full version of the original paper by Cohen et al.~\cite{cohen_certified_2019}. Using these results, an algorithm for certifying the predicted class from an input $x$ can be constructed as follows.

% \begin{algorithm}
% 	\caption{Certification using Randomised Smoothing \cite{cohen_certified_2019}} 
% 	\label{list:certify}
% 	\begin{algorithmic}[1]
% 		\Function{certify}{$f, \sigma, x, n_0, n, \alpha$}
% 		\State \verb|counts0| $\gets$ SampleUnderNoise($f, x, n_0, \sigma$)
% 		\State $\hat{c}_A \gets $ top index in \verb|counts0|
% 		\State \verb|counts| $\gets$ SampleUnderNoise($f, x, n, \sigma$)
% 		\State $\underline{p_A}$ $\gets$ LowerConfBound($\verb|counts|[\hat{c}_A], n, 1 - \alpha$)
% 		\If{$\underline{p_A} > \frac{1}{2}$}
% 		\State \Return{prediction $\hat{c}_A$ and radius $\sigma \Phi^{-1}(\underline{p_A})$}
% 		\Else
% 		\State \Return abstain
% 		\EndIf
% 		\EndFunction
% 	\end{algorithmic} 
% \end{algorithm}

For a base classifier $f$ and input $x$, $n_0$ counts of noise are sampled from $\mathcal{N}(0, \sigma^2I)$, and the modal class $c_A$ is selected as the target class. In order to perform the certification, the output classes are sampled under noise from $f$, $n$ times. A lower confidence bound for $p_A$ with confidence $\alpha$ is obtained using the Clopper-Pearson method \cite{clopper_use_1934}. $\overline{p_B}$ is estimated as $1 - \underline{p_A}$. Finally, the radius of certification can be computed using \Cref{eqn:radius}. While $n_0$ can be relatively small to determine $c_A$ effectively, $n$ needs to be quite large. Approximately $10^5$ samples are required to certify a radius of $4\sigma$ with 99.9\% confidence \cite{cohen_certified_2019}.

\subsection{Randomness in (Adversarial) ML}

Randomness is heavily used in 
machine learning. Stochastic Gradient Descent, the randomised optimisation algorithm that arguably enabled modern deep learning, relies on randomness for data sampling~\cite{robbins1951stochastic}; randomised dropout improves generalisation~\cite{gal2016dropout}; active learning approaches rely on biased samplers to enable faster learning~\cite{kirsch2019batchbald}; randomised transformations are used to introduce trained invariance~\cite{lyle2020benefits}; while randomised sampling of data can lead to improved privacy\cite{balle2018privacy}, and even bound privacy attacks~\cite{thudi2022bounding}. 

Adversarial ML uses randomness in both attack and defense. Adversarial examples benefit significantly from random starts~\cite{croce2020reliable}, while many ML system enginners advocate defences based on random pre-processing where inputs are stochastically transformed before inference~\cite{raff2019bart,gao2022on,ahmed2022kws}. We are aware of only two attacks so far that depend on exploiting randomness: the batch reordering attack~\cite{NEURIPS2021_959ab9a0} and the randomised augmentation attack~\cite{rance2022augmentation}, where the order of data and the randomness of the augmentation are manipulated respectively to introduce backdoors into a target model. 

Instances of randomness failure are not uncommon, especially in the context of differential privacy, where they have occasionally led to safety and security issues. A noteworthy example dates back to 2012 when Mironov discovered that the textbook implementations of floating point numbers for Laplace noise sampling resulted in an inaccurate distribution of sampled values~\cite{mironov2012significance}. This led to a violation of privacy guarantees and highlighted the significance of the problem. More recently, a timing side-channel was discovered in the implementation of randomness samplers, once more violating privacy~\cite{jiankai2022timingsidechannelindp}. Finally, randomness plagues reproducibility in ML -- model training is highly stochastic~\cite{zhuang2021randomness}, which is in tension with repeatable model training~\cite{jia2021pol}.

\section{Prior on Randomness}
\label{sec:related:rng}

Monte Carlo algorithms like the one described in \Cref{sec:related:randomisedsmooth} require random sampling. This immediately leads an inquisitive mind to question, ``what is a random number, and how can one be generated?'' The mathematical definition of a random sequence of numbers has been the subject of much debate since before computer science came into existence. For an in-depth discussion of random sequences and their limitations, the reader is referred to chapter 3.5 of The Art of Computer Programming by Knuth \cite{knuth_chapter_1997}. What follows is a summary of some points relevant to this paper.

\subsection{Random Number Generators}
While many distributions can be sampled to generate random sequences, in the context of computer programming, generally the uniform distribution $\mathcal{U}(a, b)$ is used, \ie~every number between $a$ and $b$ has an equal probability of selection. As discussed later in~\Cref{sec:sampling-normal}, the standard uniform distribution ($\mathcal{U}(0, 1)$) can be transformed into any other random distribution and is therefore a natural starting point for random number generators (\texttt{RNG}). There are quite a few sources of entropy that a computer can use to generate random numbers such as the timing of keystrokes or mouse movements \cite{ferguson_chapter_2010}. For example, consider an \texttt{RNG} relying on timing of keystrokes as the source of entropy. Generating one random bit from this \texttt{RNG} can be modelled as an experiment phrased as ``Is the time taken between two keystrokes in milliseconds an even number?". The outcome of this experiment can either be 0 or 1, thereby generating a random bit. The function that assigns a real value to each outcome of a random experiment such as the one described here is called a \textit{random variable} \cite{ross_probability_2022}. The \textit{distribution function }of a 1D random variable $X$ over the real line can be defined as 

\begin{equation}
	\label{eqn:cdf}
	F(x) = \mathbb{P}\{X \leq x\} = \mathbb{P}\{X \in (-\infty, x]\}.
\end{equation}

% The entropy of a random variable is a measure of its randomness \cite{shannon_mathematical_1948} and can be defined as

% \begin{equation}
% 	H(X) := - \sum_x p(x) \log (p(x)).
% \end{equation}

At the dawn of computing, a lot of research effort was spent on developing efficient ways of generating random numbers. There were mechanical devices such as ERNIE \cite{kendall_randomness_1938} which was used to generate random numbers for an investment lottery and could possibly be attached to computers. Modern CPUs now often feature built-in hardware \texttt{RNG}s that use miniscule natural fluctuations in current to generate random bits \cite{taylor_behind_2011}. However, the limitations of using mechanical \texttt{RNG}s in the 1950s are applicable to these modern hardware \texttt{RNG}s as well. First, a hardware \texttt{RNG} makes it difficult to reproduce the functioning of a randomised program to check if it is behaving as expected. There is also a possibility that the machine will fail in ways that would be difficult to detect by a program using it \cite{knuth_chapter_1997}. Another problem is that it is difficult to judge the level of entropy of an \texttt{RNG} relying on real-world properties \cite{ferguson_chapter_2010}. Finally, of special interest to the problem being tackled in this work is the fact that it is very tricky to determine the distribution function of a hardware \texttt{RNG}, as this can depend on environmental factors outside the control of the system designer.

These issues led to the development of pseudo-random number generators (\texttt{PRNG}s) which rely on deterministic calculations to generate sequences of numbers that are not truly random, yet appear to be. Von Neuman \cite{von_neumann_various_1951} proposed the first \texttt{PRNG} in 1946 using the middle-square method. \texttt{PRNG}s use the current number in the sequence or some other counter to generate the next number, but they need a starting point -- a seed -- to generate the first number or counter value. The sequence of numbers is thus a function of the seed, and two instances of the same \texttt{PRNG} will produce the same sequence if the seeds are identical\footnote{Not all \texttt{PRNG}s work like this, but this is a desirable property to ensure reproducibility. Cryptographic \texttt{PRNG}s typically combine hardware and pseudorandom components in such a way that both have to fail to make key material easily predictable by an opponent.}. The following sections present the process of generating random numbers and transforming them into samples from the normal distribution. This is followed by an overview of statistical tests to determine the quality of random numbers produced by a \texttt{PRNG}, and finally a discussion of possible attacks.

\subsection{The Linear Congruential Method and PCG64}
\label{sec:related:pcg64}

Most popular random number generators use the linear congruential method, first introduced by Lehmer in 1949 \cite{lehmer_mathematical_1951}. This produces a sequence of numbers $X_1, X_2, \cdot$ as follows \cite{knuth_art_1997}:

\begin{equation}
	\label{eqn:lcm}
	X_{n+1} = (a X_n + c) \mod m,\ n \geq 0,
\end{equation}

where:
\begin{align*}
	m > 0&\text{ is the modulus},\\
	0 \leq a \leq m&\text{ is the multiplier} ,\\
	0 \leq c \leq m&\text{ is the increment, and}\\
	0 \leq X_0 \leq m&\text{ is the seed}.
\end{align*}

\Cref{eqn:lcm} can be used to generate random numbers between 0 and $m$. Different choices of $a$ and $c$ affect the performance. The permuted congruential generator (PCG) \cite{oneill_pcg_2014} is widely used and uses an adaptation of this method. The 64-bit version has a 128-bit state variable, which is advanced according to \Cref{eqn:lcm}, \ie~$X_n$ gives the $n$th state and is a 128-bit integer. A 64-bit random number can be generated from this 128-bit state as:
\begin{verbatim}
output=rotate64(state^(state>>64)),state>>122)
\end{verbatim}

First, the state is bit shifted right by 64 bits and XORed with itself to improve the randomness in high bits. Then a clockwise rotational bit shift of $r$ bits is applied to the lower 64 bits of the resulting value, where the value of $r$ (between 0 and $2^6$) comes from the 6 leftmost bits of the \texttt{PRNG} state\footnote{Typecasting the 128-bit state to \texttt{uint64\_t} in C retains the lower 64 bits and discards the top 64 bits.}. For more details on the design choices and an empirical analysis of its performance, the reader is referred to O'Neill \cite{oneill_pcg_2014}. The PCG64 generator is the default choice for the Numpy library \cite{harris_array_2020} and is the target of the attack presented in this work. Other popular random number generators include the Mersenne Twister \cite{haramoto_fast_2008, lecuyer_f2-linear_2009}, SFC64 (Small Fast Chaotic) \cite{doty-humphrey_practrand_2014} and Philox \cite{salmon_parallel_2011}. Pytorch uses Philox, which is a counter passed through a block cipher. As it was designed to achieve high performance in HPC applications, the cipher was deliberately weakened \cite{salmon_parallel_2011}.

\subsection{Sampling from a Normal Distribution}
\label{sec:sampling-normal}

The numbers generated by PCG64 can be assumed to be sampled from the discrete uniform distribution $\mathcal{U}(0, 2^{64})$. However, most randomised algorithms need this to be transformed into a different distribution, usually $\mathcal{U}(0, 1)$ or $\mathcal{N}(0, 1)$, which can then be transformed into any uniform or normal distribution trivially. Transforming discrete $\mathcal{U}(0, 2^{64})$ to continuous $\mathcal{U}(0, 1)$ can be done by simply discarding the first 11 bits of the random number\footnote{These bits contain information about the exponent in 64-bit floating point numbers \cite{noauthor_ieee_2019}}, typecasting as double, and dividing by $2^{53}$ to effectively shift the decimal point from after the least significant bit to before the most significant bit. 

\begin{figure}
    \centering
    \includegraphics[width=\linewidth]{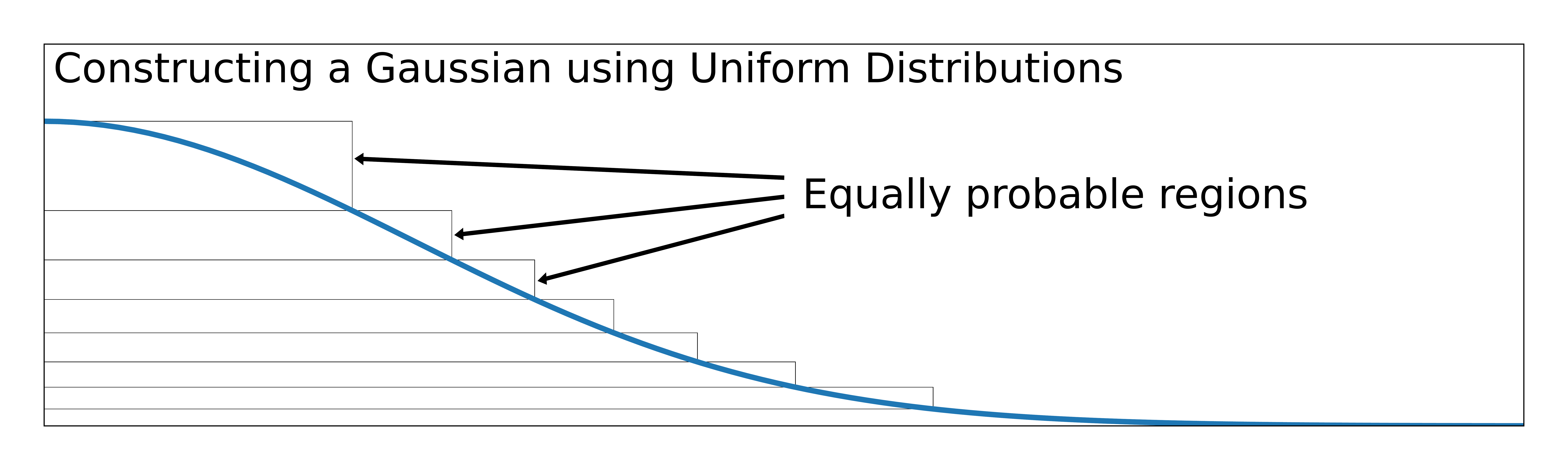}
    \caption{Sampling from a Gaussian distribution constructed using Uniform distributions with the Ziggurat Method~\cite{marsaglia_ziggurat_2000}.}
    \label{fig:ziggurat}
\end{figure}

The Ziggurat method \cite{marsaglia_ziggurat_2000} is state-of-the-art when it comes to sampling from decreasing densities such as the normal distribution. At a high level, this involves dividing the distribution into smaller pieces, choosing one of these pieces randomly, and then sampling from it. The smaller pieces into which the distribution is divided are chosen such that the areas under the density curve are equal, \ie~each piece is equally likely (shown in \Cref{fig:ziggurat}). The lower 8 bits (index bits) are used to select one of these pieces at random, the next bit (sign bit) is used to randomly assign a sign, and finally, the remaining bits (distribution bits) are used to sample a float from the uniform distribution corresponding to the rectangular area selected using the index bits.

\subsection{Attacks, Dual\_EC\_DRBG and Bullrun}
\label{sec:rng:dualec}

Traditional attacks on pseudo-random number generators have focused on determining the next number to be generated in a sequence; or more generally, given some of the numbers in a sequence, deducing some others. If the generator is used to produce cryptographic keys and also other random values that may become visible to the attacker, such a deduction may compromise an encrypted message. Such attacks aim to determine the state of the \texttt{PRNG} being used by the victim. Since most \texttt{PRNG}s are deterministic, a known state can be used to determine future numbers in the sequence \cite{ferguson_chapter_2010}.  Another class of attacks involves introducing a backdoor in the \texttt{PRNG} to make future outputs more predictable. The NSA pushed NIST to include EC\_DRBG in the ANSI X9.82 standard for random number generators when the standard was first developed at the start of the 21st century. The \texttt{PRNG} featured in the first draft of the standard, which was published in 2004. The research community expressed concerns about EC\_DRBG not being cryptographically sound by 2005~\cite{gjosteen_comments_2006} -- before the first official version of the standard was published~\cite{barker_recommendation_2006} -- leading to conjecture of an NSA backdoor~\cite{schneier_did_2007}. Despite these concerns, EC-DBRG was adopted as the default random number generator by RSA in their BSAFE cryptography library~\cite{me2007rsa}. The Snowden leaks in 2013 confirmed the existence of project Bullrun, which aimed ``to covertly introduce weaknesses into the encryption standards followed by hardware and software developers worldwide" and successfully injected a backdoor into the default \texttt{PRNG} used for all cryptographic encryption between 2006 and 2013. This is a useful reminder of the tactics, tools, and procedures that may be used by a capable motivated adversary to carry out a backdoor attack against pseudo-random number generators.

%% file: sections/methodology.tex
\section{Methodology}

\subsection{Threat Model}
\label{sec:threat-model}

It is assumed that the victim is attempting to use ML to get predictions in a safety-sensitive or security-sensitive setting. Therefore it is important to accurately gauge the robustness of every prediction, and we assume that~\RS~is being employed for this purpose. By definition, the probability that an adversarial example can be found within the certified radius obtained from randomised smoothing is low: less than 0.1\% if $\alpha$ is set to $0.001$ as suggested by Cohen et al.~\cite{cohen_certified_2019}. The knowledge that the victim is employing randomised smoothing can in itself be very useful to an adversary. This was demonstrated by Cullen et al.~\cite{cullen_exploiting_2023}, who used this information to only search for adversarial examples with $l^2$ distance greater than the certified radius, achieving better than the state-of-the-art success rate at finding adversarial examples against ML models using randomised smoothing. Furthermore, in their evaluation of randomised smoothing, Cohen et al.~\cite{cohen_certified_2019} found that the probability of finding adversarial examples increases rapidly as the upper bound on the $l^2$ norm of the adversarial example set by the adversary increases beyond the certified radius, $R$. For the ImageNet dataset~\cite{deng_imagenet_2009}, they found that the probability of finding an adversarial example against a smoothing classifier is 0.17 at an upper bound of $1.5R$ and 0.53 at $2R$. Hence, confidence serves a good proxy for reliability of prediction, and any compromise of~\RS~will decrease its trustworthiness. 

The objectives of an attacker attempting the class of attacks presented in this paper are twofold, depending on whether the certified radius obtained from randomised smoothing is being manipulated to be higher or lower. A higher certified radius can make the victim believe that the prediction is more robust for a given input than it actually is, so that the victim ignores adversarial examples within the spoofed radius. And certification is costly, requiring approximately $10^5$ inferences from the smoothed model to certify a radius of $4\sigma$. Reducing the certified radius can make the victim waste time and compute, forcing them to generate more predictions under noise to obtain the desired radius, providing a service-denial attack. 

In this paper we limit our adversary access to the underlying random number generator used by the victim to perform randomised smoothing. While the setting may seem unrealistic, an attack with such access happened in the past. The story of the Dual\_EC\_DBRG \texttt{PRNG}, which the NSA used to compromise RSA libraries, was told in~\Cref{sec:rng:dualec}. 

The setup used by the victim to perform training and certification can be quite complicated, leaving multiple avenues for an attacker to gain access and manipulate the noise. Modern practices such as ML-as-a-service, and outsourcing of training and data generation to third-parties, opened a Pandora's box. The attacks discussed here focus on modifying the noise distribution, first by using a different noise function directly and later by modifying the bitstream generated by an underlying pseudo-random number generator (\texttt{PRNG}). Where the victim has little or no control over the software and hardware used for training and certification, there are even more attack vectors. The objective of the attacks we present in the following sections is to spoof the certified radius while escaping detection by the victim -- whether by analysing the performance of the model or by using statistical tests (discussed in~\Cref{sec:method:defence}). This class of attacks can be carried out in a traditional way by exploiting the hardware or software layers. In addition, with recent developments in denoising diffusion models \cite{ho_denoising_2020}, they can be carried out in an ML-as-a-Service scenario too. If the victim submits inputs to a cloud API for prediction and certification, an attacker can effectively remove the noise introduced by the victim and replace it. 

% A visual representation of the attack is presented in~\Cref{fig:threatmodel}.

% \begin{figure*}[!ht]
% 	\centering
% 	\includegraphics[width=\linewidth]{figures/infographics/smoothness.png}
% 	\caption{Threat Model}
% 	\label{fig:threatmodel}
% \end{figure*}

\subsection{Naive noise distribution replacement}

In order to check the feasibility of the attacks described in the previous section, an initial test was done by explicitly changing the distribution that is sampled during certification. The GitHub repository released by Cohen et al.\footnote{\url{https://github.com/Xzh0u/randomized-smoothing}}~was used as a starting point. Cohen et al.~\cite{cohen_certified_2019} reported that the variance of distribution that additive noise is sampled from controls the trade-off between accuracy and robustness. A lower value of $\sigma$ can be used to certify smaller radii but with a high degree of accuracy, whereas a higher value of $\sigma$ is needed to certify large radii, resulting in lower accuracy. $\sigma = 0.25$ was found to achieve an acceptable trade-off between certification radius and accuracy on the CIFAR 10 \cite{krizhevsky_learning_2009} dataset, which was used to run all the experiments in this paper. Using a base classifier trained with additive noise sampled from $\mathcal{N}(0, 0.25^2)$, the naive attack swaps the noise distribution for one of the following when certification is performed: Laplace distribution, $\mathcal{L}(0, 0.25)$; half-normal distribution, $|\mathcal{N}(0, 0.25^2)|$; uniform distribution, $\mathcal{U}(-0.25, 0.25)$; and Bernoulli distribution, $\mathcal{B}(0.5)$. This naive attack was used to demonstrate the feasibility of this class of attacks on \texttt{PRNG}s. The next section will describe a more sophisticated attack.

\myexample{Attack Scenario Example}{
	\begin{description}
		\item \textbf{Example Setting: } Detection of an enemy tank is being performed from a satellite image. To make sure that the prediction is not a spurious correlation and to counteract camouflage paint, \RS~is used. \texttt{PRNG} generates the noise for~\RS. Randomness can come from the inference platform \ie~the cloud hardware or ML-as-a-Service API; alternatively the user can apply noise themselves\footnote{Note that this may result in quantisation artefacts, \eg~pixel with value 245 noised to 235.3, gets quantised to 236, and potentially change performance of~\RS. }.
		\item \textbf{A Normal User: } A user attempting to certify the robust $l^2$ radius that does not contain adversarial examples around an input in the above setting. Here, our user may be an analyst that aims to find enemy tanks to target by a missile system.  
		
        \item \textbf{Attacker: } An adversary with the objective of manipulating the certified radius obtained by the user. An attacker may increase it to make the prediction appear more robust than it actually is, \ie~convince an the user that tank is present at a given location, or decrease it to make the user uncertain about their prediction or even abstain from it, \ie~force them to not notice a tank at a given location.

		\item \textbf{Defender: } The defender's goal is to detect the presence of a randomness-based adversary to determine if the noise function used for certification deviates significantly from white noise or not.
	\end{description}
}

\subsection{Bit-flipping \texttt{PRNG} attacker}
\label{sec:method:PRNG}

The objective of the bit-flipping \texttt{PRNG} attacker is to modify the stream of bits being generated by the random number generator to alter the distribution when normal variates are sampled. An overview of the algorithm used to transform random integers to floats in the standard normal distribution is presented in \Cref{sec:sampling-normal}. Taking a 64-bit random integer, the rightmost 8 bits determine the rectangle in which the normally distributed random number will fall. These are called the \textit{index bits}. The 9th bit is the \textit{sign bit}. Finally, the remaining 55 bits are transformed into a floating point number falling within the limits of the uniformly distributed rectangle chosen by the index bits. These are called the \textit{distribution bits}. The following attacks modify one of these three categories of bits to alter the resulting distribution. Altering a distribution can be done by either introducing kurtosis or skewness. Kurtosis is a measure of the tailedness whereas skewness is a measure of the asymmetry of the distribution around the mean.

All attacks described in this section were performed on the PCG64 random number generator \cite{oneill_pcg_2014} in the \texttt{numpy} library \cite{harris_array_2020}, which was then used to sample noise when performing randomised smoothing. PCG64 is a NIST-certified pseudo-random number generator. For an overview of how it works, the reader is referred to \Cref{sec:related:rng}. The following attack was performed by modifying the \verb|next64| function, which is used to generate a random 64-bit integer, in the \verb|pcg64.h| file. The original version of the function is shown in \Cref{lst:pcg-original}.

\subsubsection{Negative Kurtosis Attack}

The uniform distribution increased the relative certified radius the most and it builds the intuition behind the negative kurtosis attack. By modifying the distribution bits, so that they are no longer uniformly distributed, but positively skewed, the kurtosis in the normal distribution is reduced, as there are fewer random numbers near the mean and more towards the tails. The modification to the PCG64 code is shown in~\Cref{lst:pcg-kurtosis}. Consider a uniformly distributed random variable, $X = \mathcal{U}(0, 1)$. The probability density function (pdf) for $X$ is:

\begin{equation}
	p(x) = \begin{cases}
		1 & 0 < x < 1\\
		0 & \text{otherwise}
	\end{cases}.
\end{equation} 

This can be skewed to a new random variable $X'$ by altering the pdf as follows (for $b$ in \Cref{eq:b_def}):

\begin{equation}
	p'(x) = \begin{cases}
		ax + b & 0 < x < 1 \\
		0 & \text{otherwise}
	\end{cases}.
\end{equation}

In order to transform $X$ to $X'$, first, the cumulative distribution function, $F'$ of $X'$ must be derived. 

\begin{align}
	F'(x) &= \int_0^x p'(y) \ dy = \int_0^x ay + b\ dy \\
	&= \left[\frac{ay^2}{2} + by\right]_0^x = \frac{ax^2}{2} + bx.
\end{align}

Since $F'(x): (0, 1) \rightarrow (0, 1)$, and $0 < X < 1$, $X$ can be transformed to $X'$ by inverting $F'$. For $x \in X$, the corresponding $x' \in X'$ can be computed as follows:

\begin{align}
	x' = F'(x)
 % \frac{-b + \sqrt{b^2 - 4\frac{ax}{2}}}{2 \frac{a}{2}}\\
	= \frac{-b + \sqrt{b^2 - 2ax}}{a}.
\end{align}

The degree of skewness in $X'$ can be controlled by changing $a$. Since $F(1) = 1$ from the definition of $X'$, 

\begin{equation}\label{eq:b_def}
	b = 1 - \frac{a}{2}.
\end{equation}

\begin{figure*}
\begin{subfigure}{0.33\linewidth}
	\centering
	\begin{subfigure}{0.49\linewidth}
		\includegraphics[width=\linewidth]{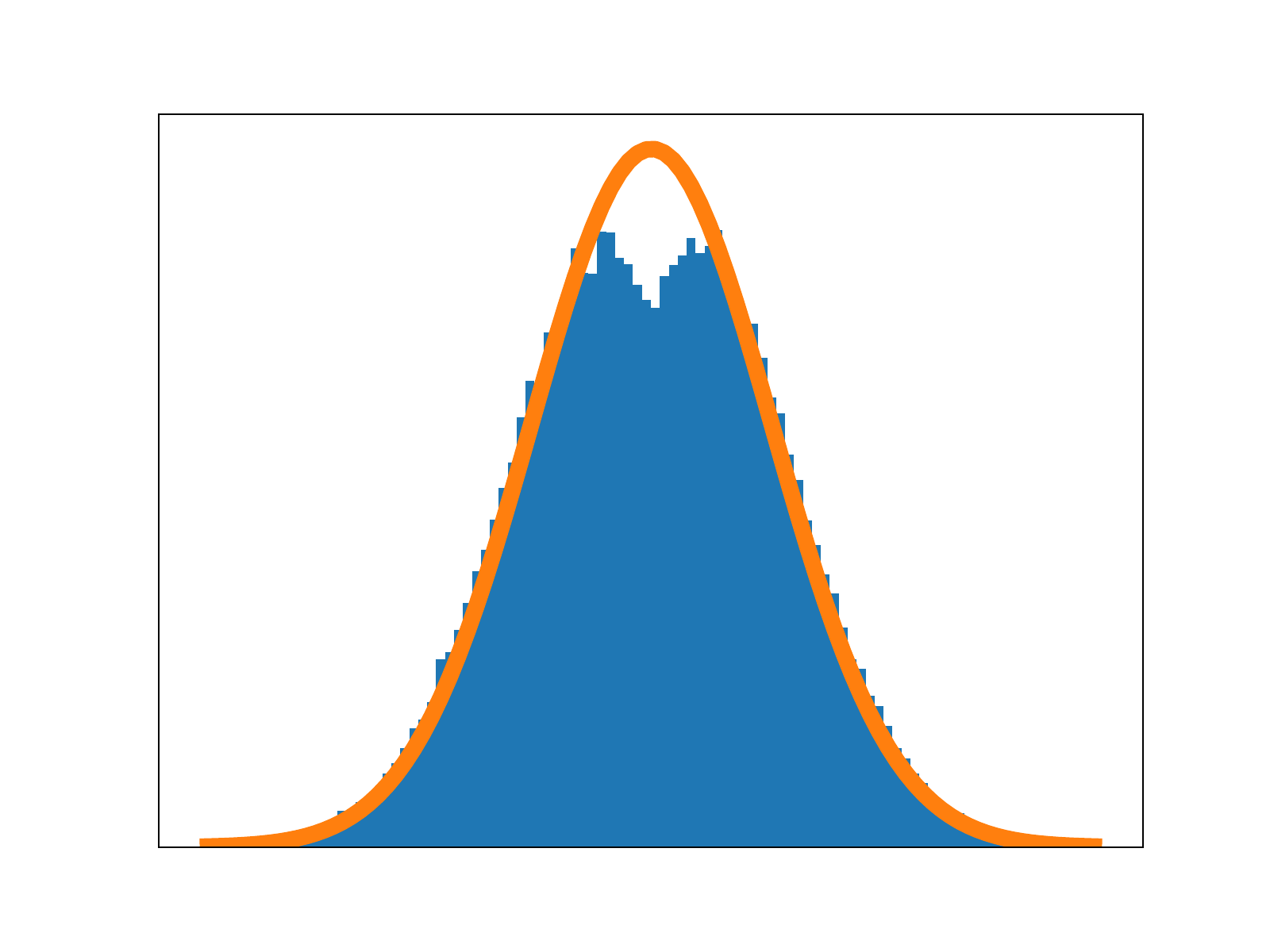}
		\caption*{$\alpha = 1$}
		\label{fig:alpha1}
	\end{subfigure}
	\begin{subfigure}{0.49\linewidth}
		\centering
		\includegraphics[width=\linewidth]{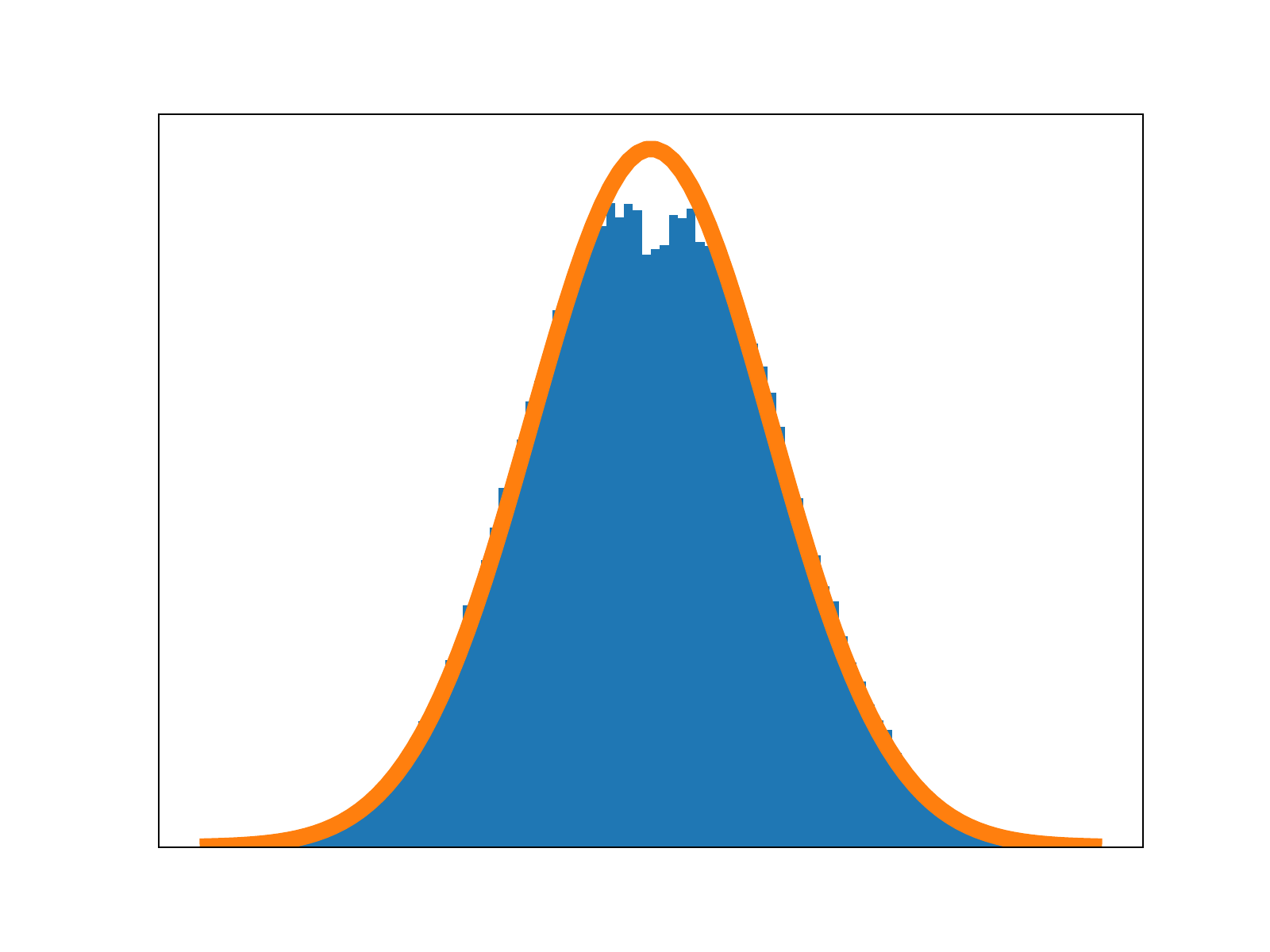}
		\caption*{$\alpha = 2$}
		\label{fig:alpha2}
	\end{subfigure}\\
	\begin{subfigure}{0.49\linewidth}
		\centering
		\includegraphics[width=\linewidth]{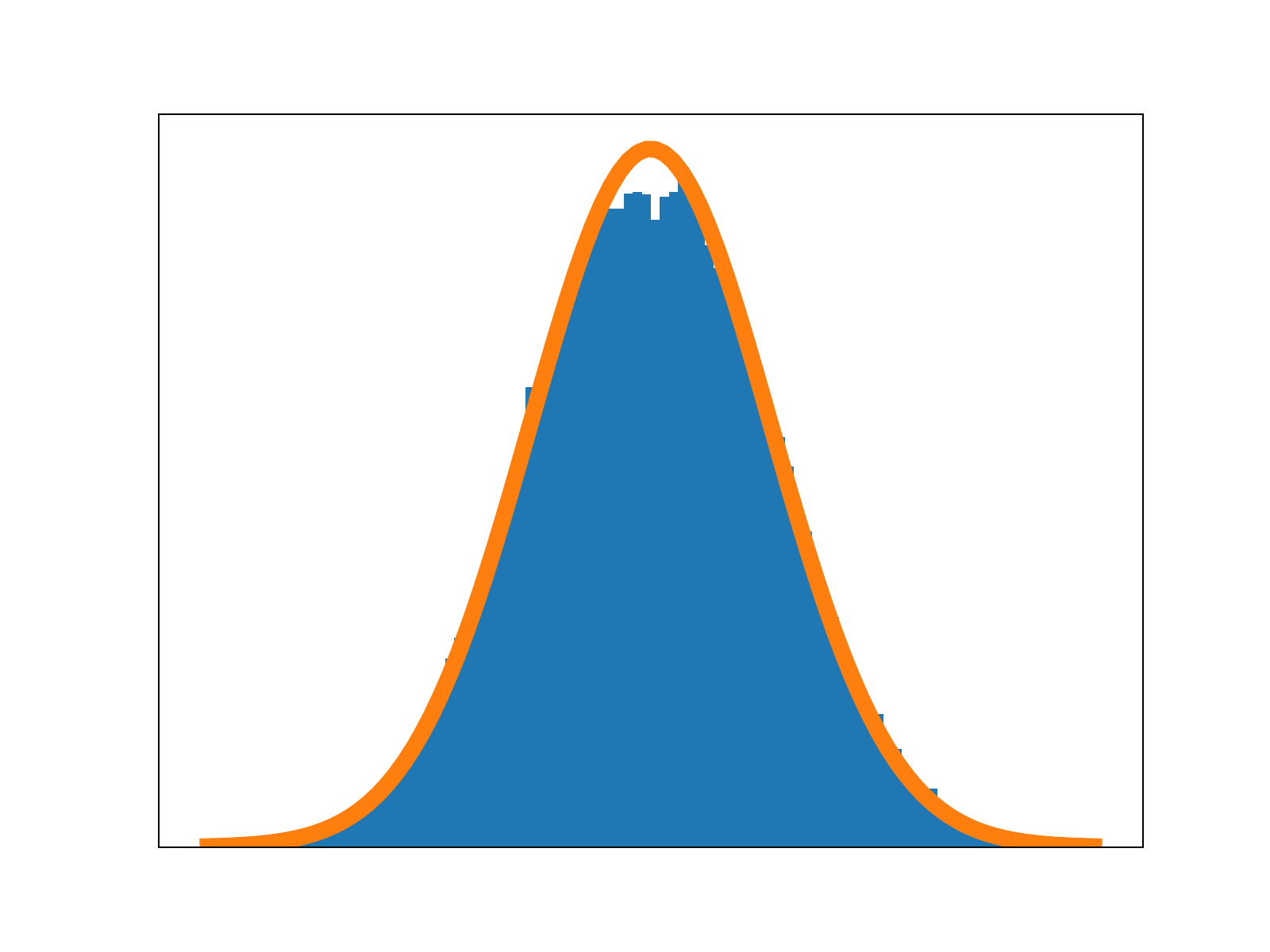}
		\caption*{$\alpha =3 $}
		\label{fig:alpha3}
	\end{subfigure}
	\begin{subfigure}{0.49\linewidth}
		\centering
		\includegraphics[width=\linewidth]{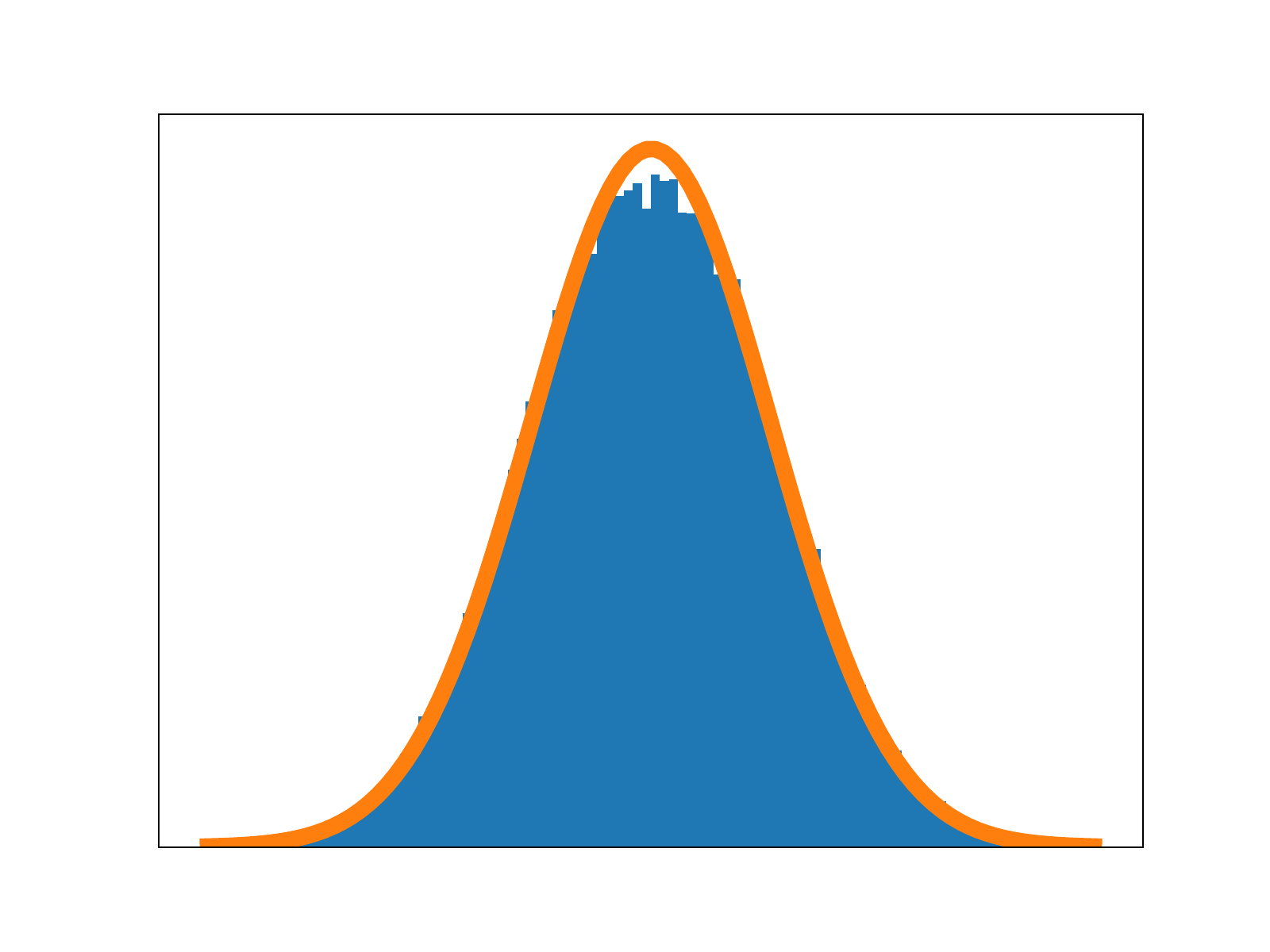}
		\caption*{$\alpha = 4$}
		\label{fig:alpha4}
	\end{subfigure}
	\caption{Negative kurtosis attack}
	\label{fig:neg-kurt}
\end{subfigure}\quad
\begin{subfigure}{0.33\linewidth}
	\centering
	\begin{subfigure}{0.49\linewidth}
		\includegraphics[width=\linewidth]{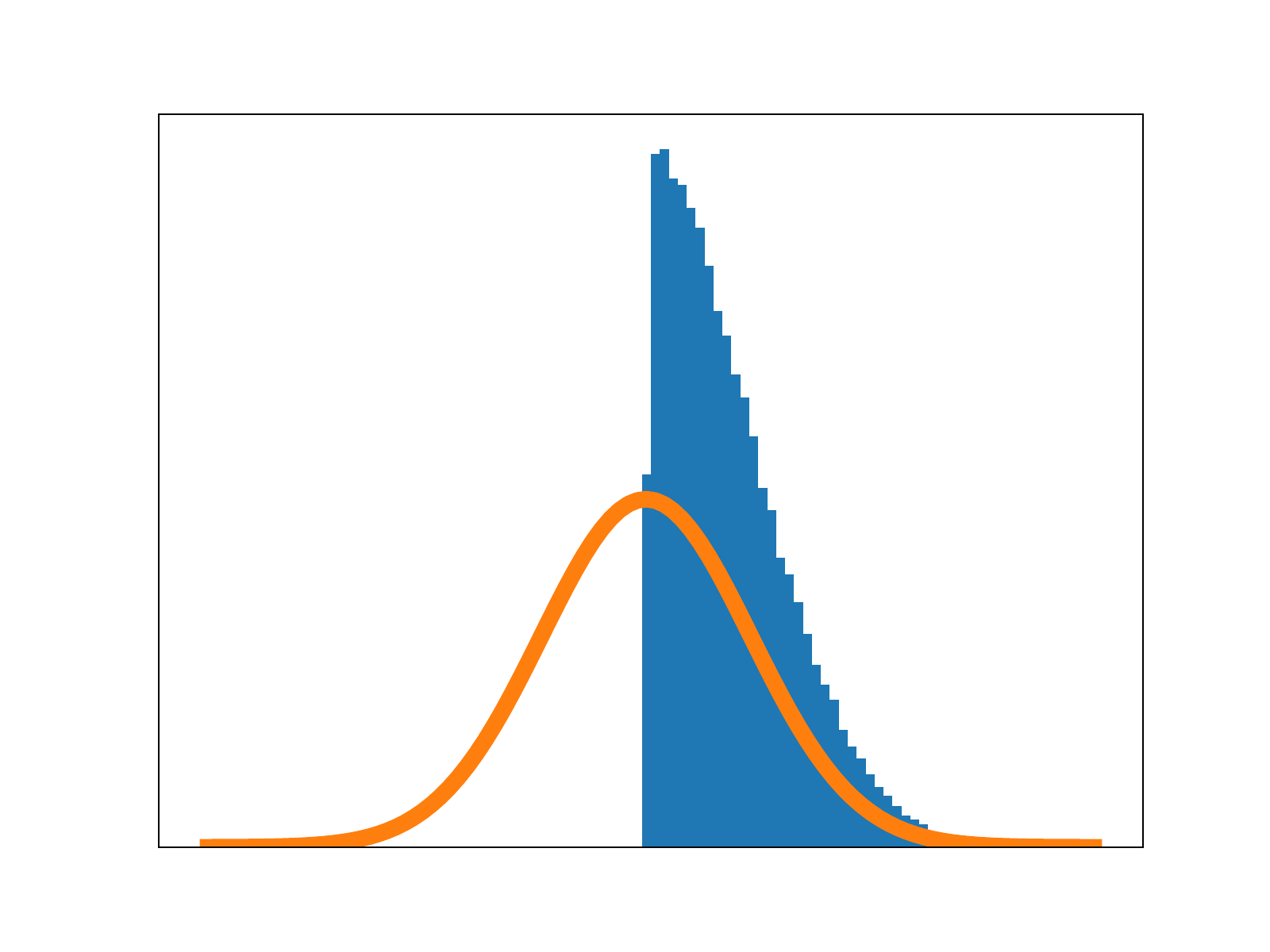}
		\caption*{$\beta = 0$}
		\label{fig:beta1}
	\end{subfigure}
	\begin{subfigure}{0.49\linewidth}
		\centering
		\includegraphics[width=\linewidth]{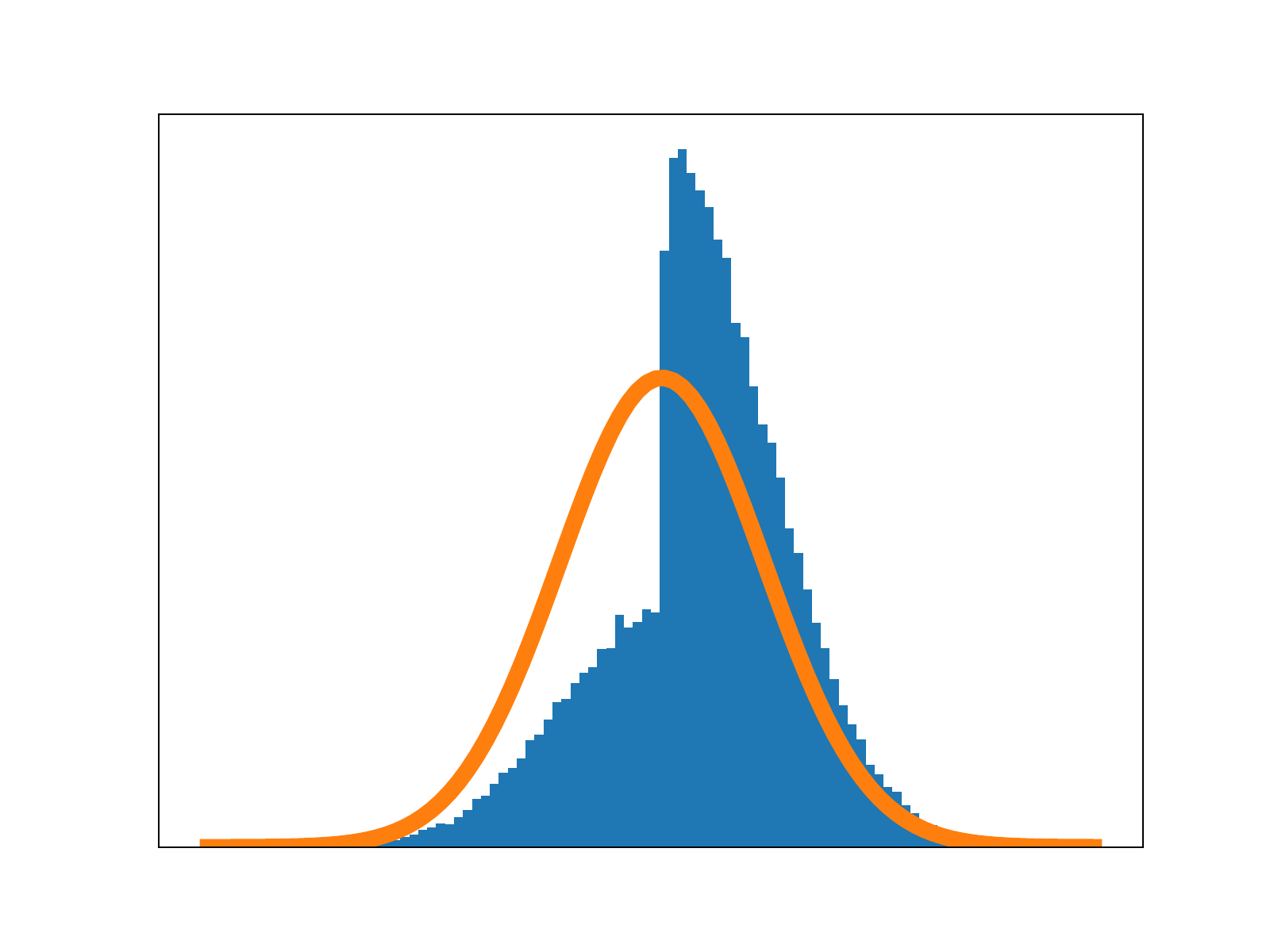}
		\caption*{$\beta = 1$}
		\label{fig:beta2}
	\end{subfigure}\\
	\begin{subfigure}{0.49\linewidth}
		\centering
		\includegraphics[width=\linewidth]{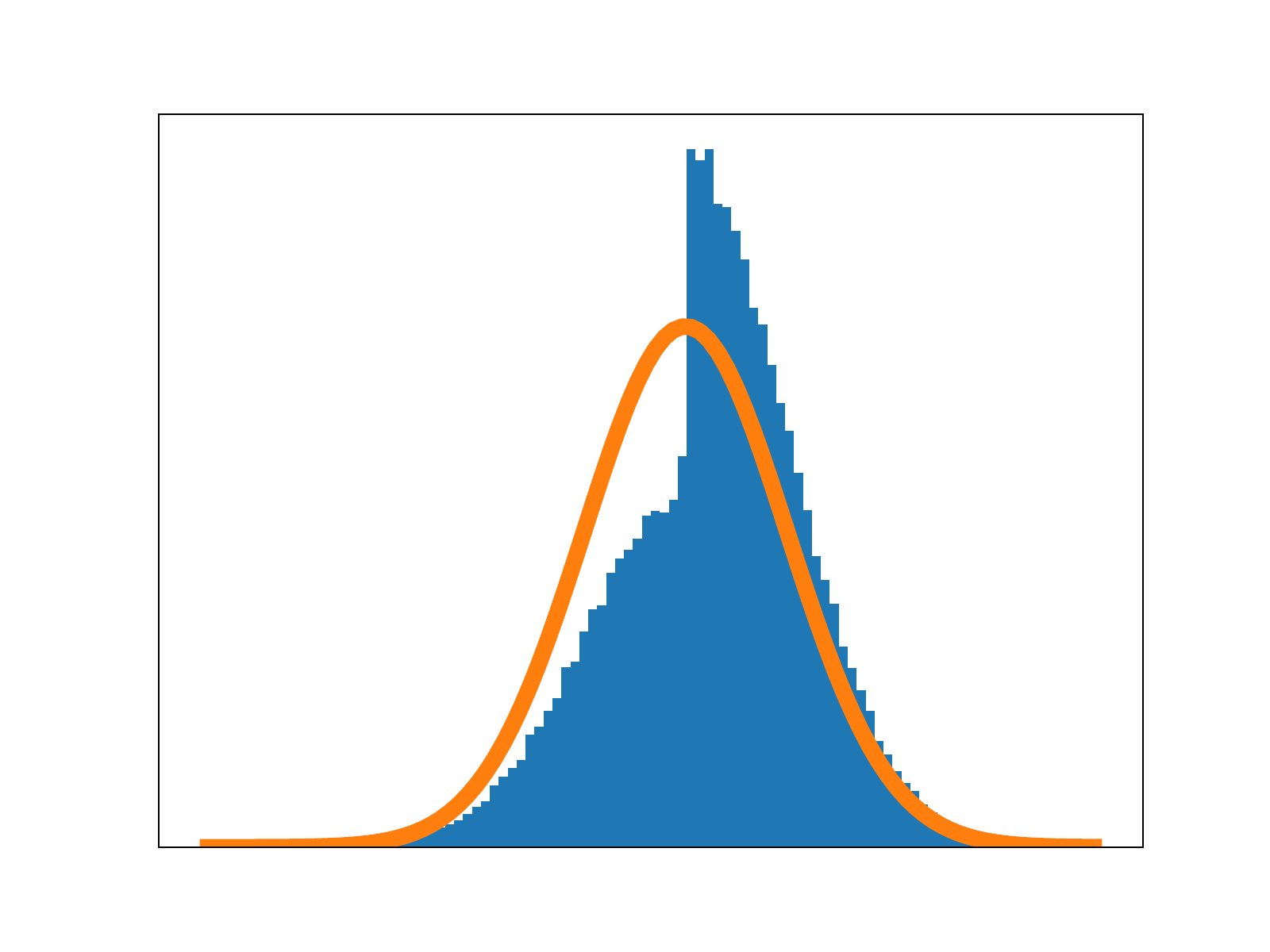}
		\caption*{$\beta =2 $}
		\label{fig:beta3}
	\end{subfigure}
	\begin{subfigure}{0.49\linewidth}
		\centering
		\includegraphics[width=\linewidth]{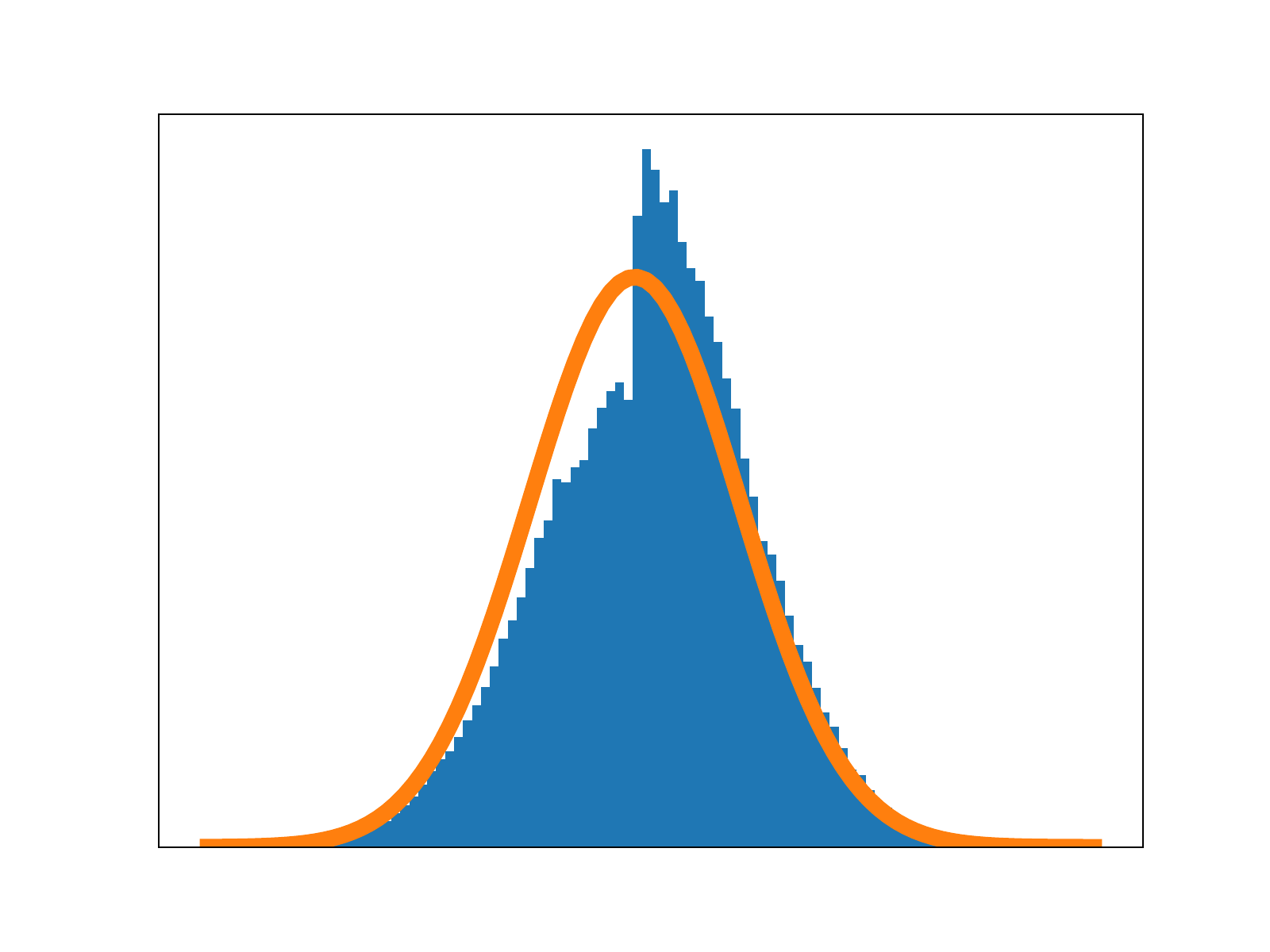}
		\caption*{$\beta = 4$}
		\label{fig:beta4}
	\end{subfigure}
	\caption{Skewness attack}\label{fig:skewness}
\end{subfigure}\quad
\begin{subfigure}{0.33\linewidth}
	\centering
	\begin{subfigure}{0.49\linewidth}
		\includegraphics[width=\linewidth]{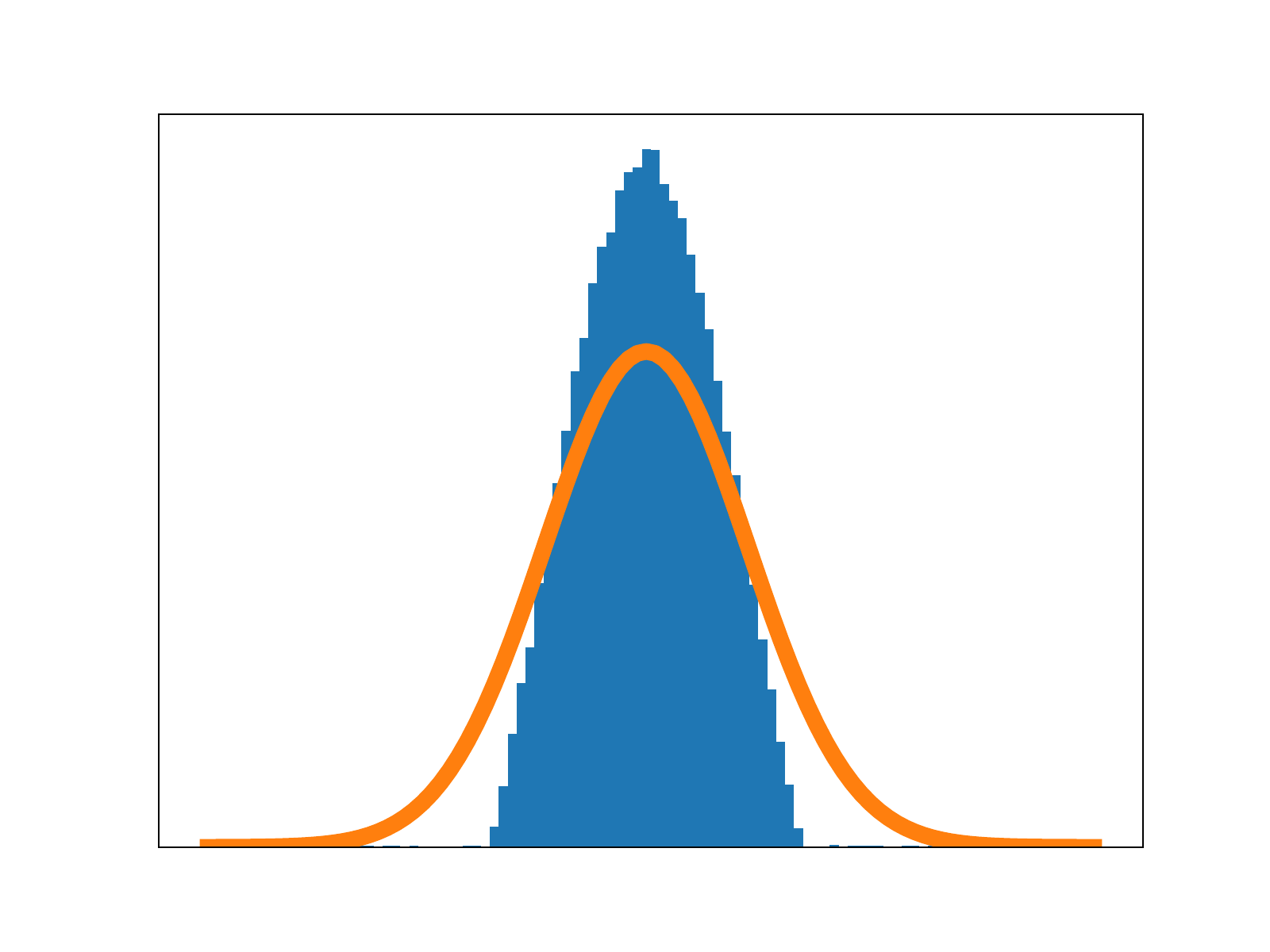}
		\caption*{$\gamma = 0$}
		\label{fig:gamma1}
	\end{subfigure}
	\begin{subfigure}{0.49\linewidth}
		\centering
		\includegraphics[width=\linewidth]{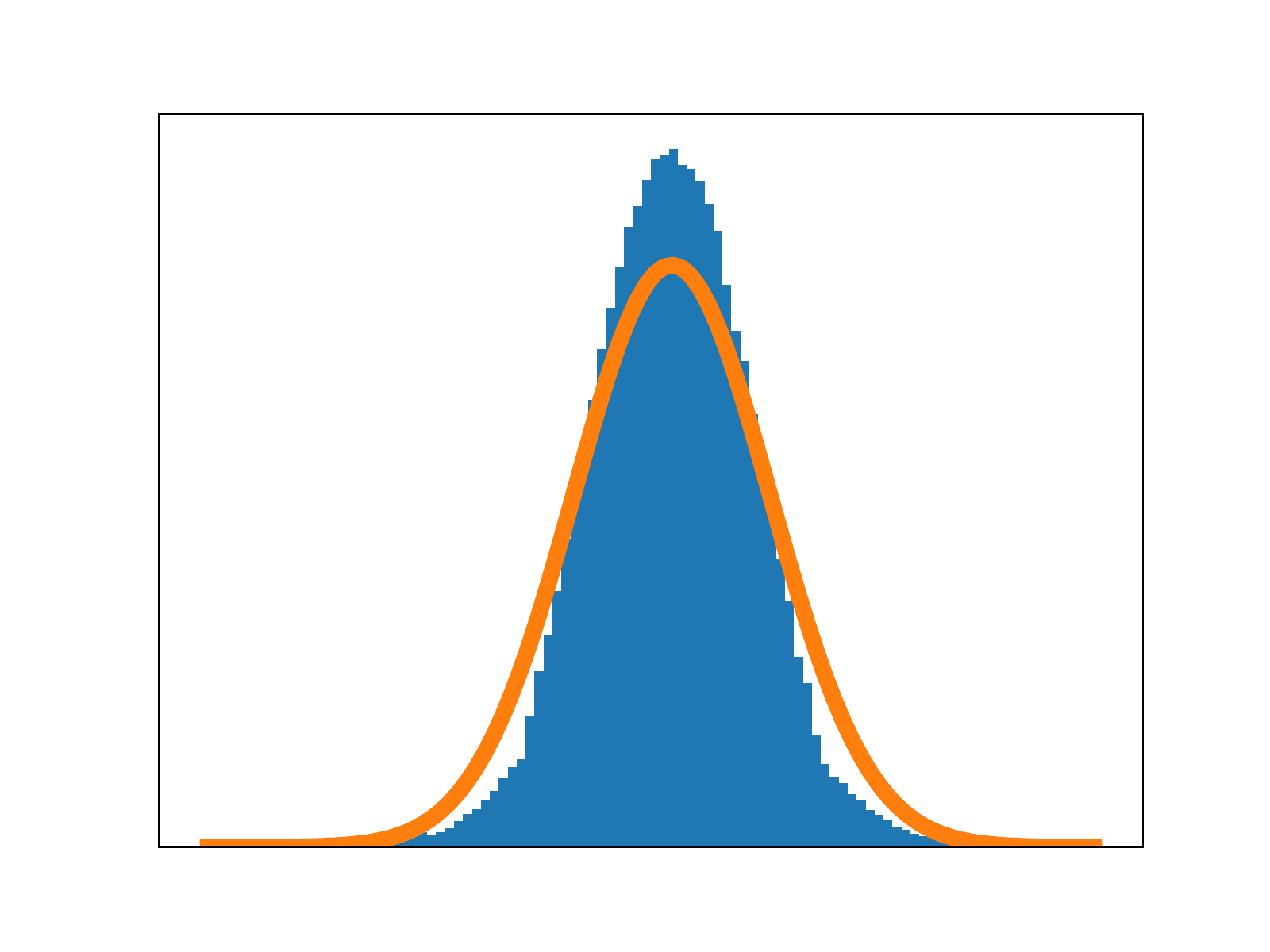}
		\caption*{$\gamma = 1$}
		\label{fig:gamma2}
	\end{subfigure}\\
	\begin{subfigure}{0.49\linewidth}
		\centering
		\includegraphics[width=\linewidth]{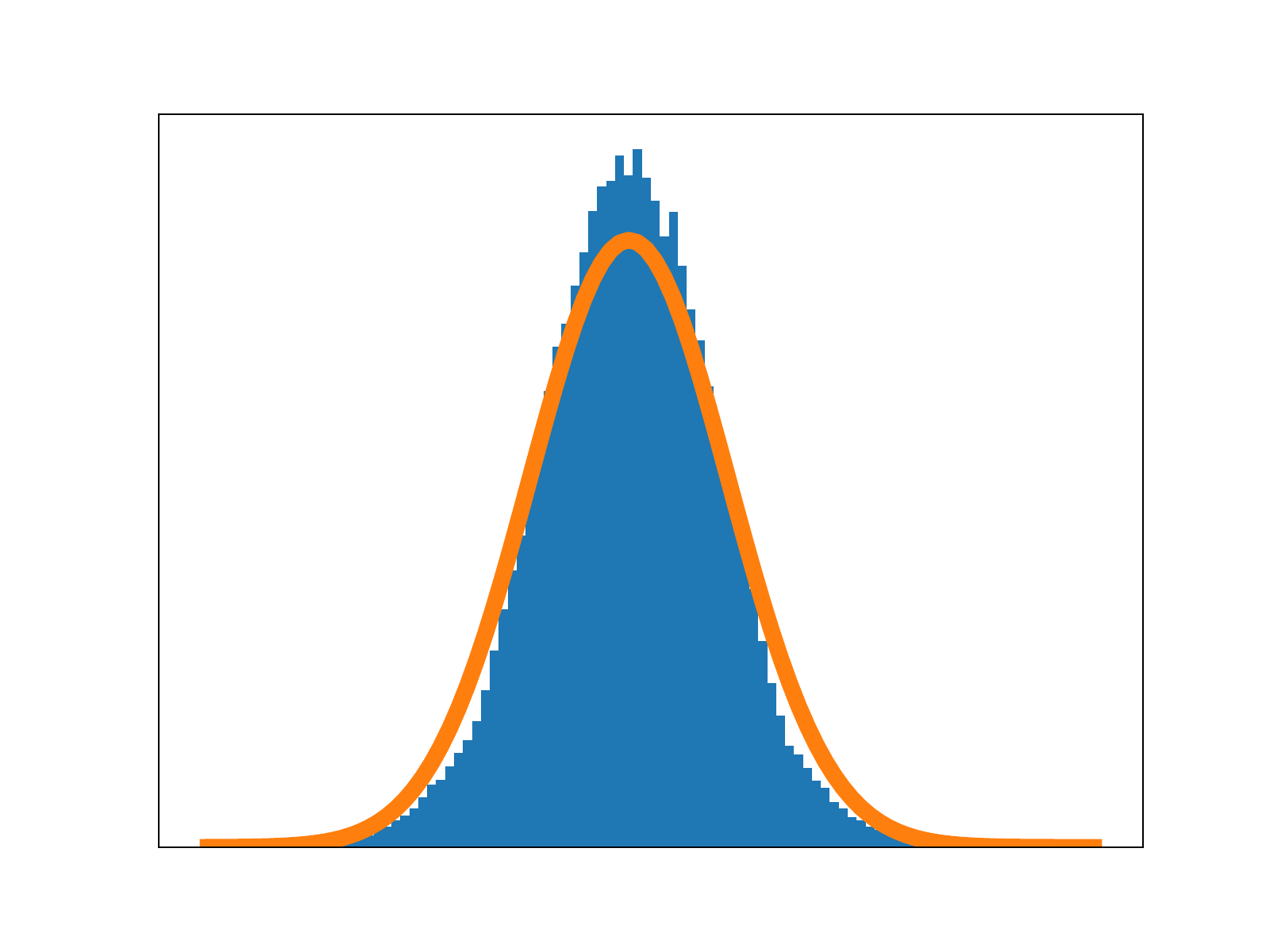}
		\caption*{$\gamma =2 $}
		\label{fig:gamma3}
	\end{subfigure}
	\begin{subfigure}{0.49\linewidth}
		\centering
		\includegraphics[width=\linewidth]{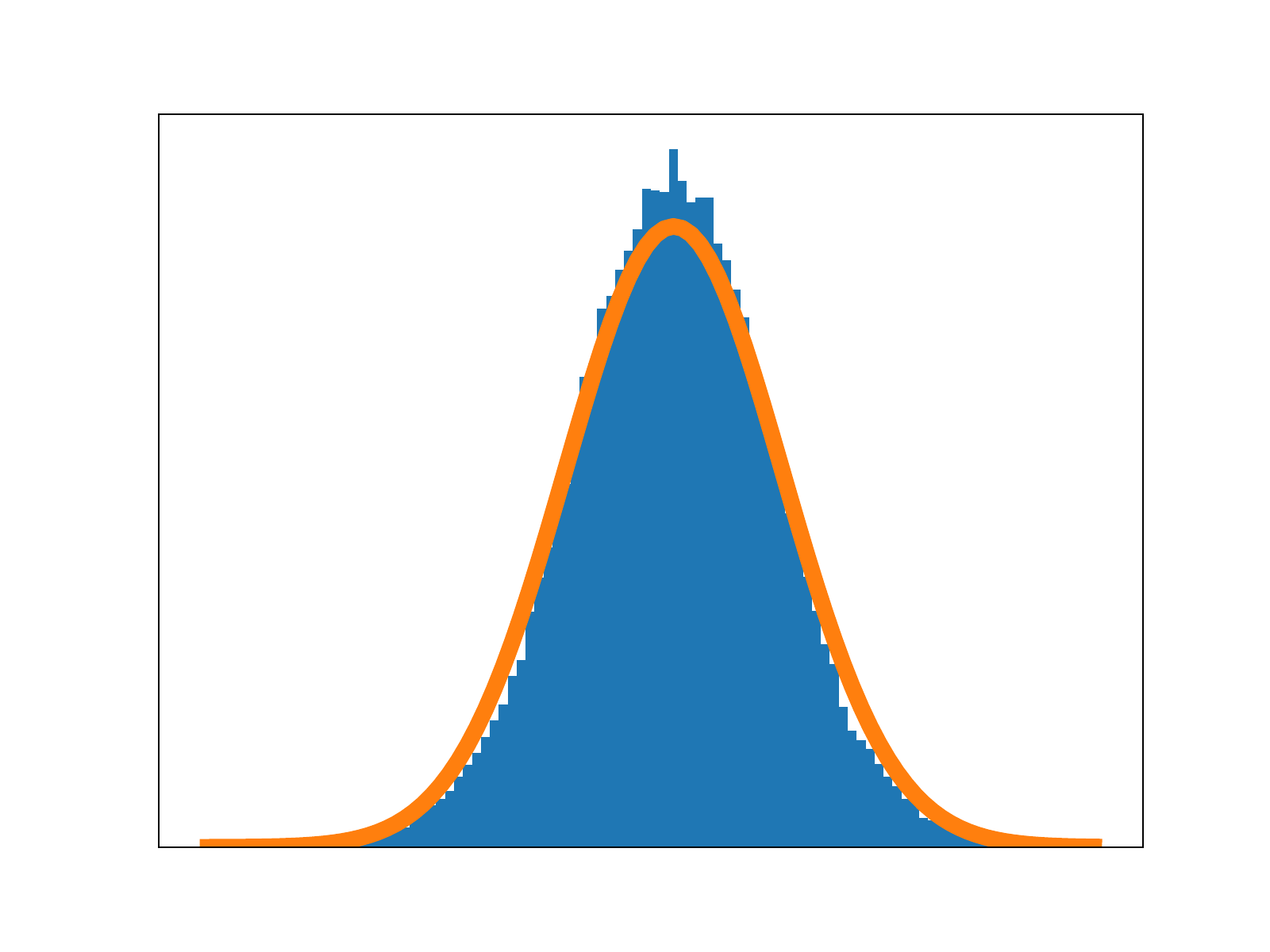}
		\caption*{$\gamma = 4$}
		\label{fig:gamma4}
	\end{subfigure}
	\caption{Positive Kurtosis attack}
	\label{fig:pos-kurt-dist}
\end{subfigure}
\caption{These figures show the sampled probability densities against the normal pdf in orange for all considered values of the tunable parameters for each of the bit-flipping \texttt{PRNG} attacks proposed. The negative-kurtosis, skewness and positive-kurtosis attacks all alter the sampled distribution in a self-explanatory manner. }
\label{fig:all_attacks_pdfs}
\end{figure*}

In~\Cref{lst:pcg-kurtosis}, the tunable parameter is $\alpha$, such that $a = 1 / \alpha$. First, the 64-bit random integer is converted to a 64-bit double-precision floating point number. Then, the transformation from $X$ to $X'$ is applied. Next, the number is converted back to a 64-bit integer. This is finally bit-shifted left by 9 bits and the least significant 9 bits are copied over from the original random integer generated by the \texttt{PRNG}. This is so that the index and sign bits remain random, and only the distribution bits are modified. A lower value of $\alpha$ results in higher skewness in the distribution bits, and more negative kurtosis in the resulting distribution. The sampled probability densities are plotted in~\Cref{fig:neg-kurt}, along with a reference curve of the probability density function of the normal distribution.

\subsubsection{Skewness Attack}

The half-normal distribution achieved a significant success rate in manipulating the certified radius to be high enough such that evasion attacks could succeed, and yet still maintain an acceptable level of performance compared to the original model. Therefore, this distribution formed the basis of the skewness attack shown in~\Cref{lst:pcg-skew}. An 8-bit integer is used as a counter, initialised to 1. Every time a random integer with the sign bit set to 1 is encountered, the counter is shifted left. When the value of the counter reaches the tunable parameter, $\beta$, the sign bit is flipped to zero, keeping the rest of the random number the same. This results in the final normally distributed random number being positive instead of negative, thereby skewing the distribution. The sampled probability distributions for $\beta = 0, 1, 2,$ and $4$ are shown in~\Cref{fig:skewness}. For $\beta = 0$, the resulting distribution is the same as $|\mathcal{N}(0, 0.25^2)|$.
\subsubsection{Positive Kurtosis Attack}

The motivation for this attack comes from the increase in certified radius observed when the normal distribution was replaced with the Laplace distribution. While the relative increase in radius achieved was not as high as the others, the attack is further motivated by the reasoning that it is more likely that the sampled prediction is the same as the modal class if the $l^2$ norm of the additive noise is lower, thereby increasing the certified radius. The modified function in PCG for this attack is shown in~\Cref{lst:pcg-zero}.

A counter is initialised to 1 and shifted left every time a random number is generated. Every time the value of the counter is equal to the tunable parameter $\gamma$, the index bits in the random number are shifted right by 1. This divides the value of the index and reduces the size of the rectangular region chosen by the algorithm used for transformation to the normal distribution. It moves the random normal numbers closer to the mean, and increases the kurtosis of the resulting distribution, with the maximum kurtosis achieved for $\gamma = 0$, and becoming closer to the standard normal distribution as $\gamma$ increases. The sampled probability densities of the distributions resulting from $\gamma = 0, 1, 2$ and $4$ are shown in \Cref{fig:pos-kurt-dist}.

\subsection{Defences}
\label{sec:method:defence}

\subsubsection{NIST Test Suite}

The National Institute of Standards and Technology (NIST) first published the SP800-22 test suite in 2001, in the aftermath of the competition in which the AES algorithm was selected. Its purpose was to ensure that random number generators used for cryptography are secure. This is done via a set of 15 statistical tests that operate on bitstreams from a random number generator, aiming to detect any deviation from a truly random sequence of bits. The reader is directed to Bassham et al. for an in-depth description of each test along with the recommended parameter values to use~\cite{bassham_iii_sp_2010}. It is important to note that this test suite is a measure of \texttt{PRNG} badness and not goodness. According to the recommendations presented by NIST, it is important that a \texttt{PRNG}'s design is based on sound mathematical principles. Due to the statistical nature of these tests, it is possible that a well-designed \texttt{PRNG} may sometimes fail a test, while a bad \texttt{PRNG} may pass some tests, as demonstrated later in this paper.

\subsubsection{Test for Normality}

We employ a number of statistical tests to detect finer fluctuations in the resulting distribution after transformation and to quantify the deviation. In the main body of the paper, we use the Shapiro-Wilk test \cite{shapiro_analysis_1965} --  empirically one of the highest-power statistical tests to determine whether a set of samples is normally distributed~\cite{razali_power_2011}. We list the performance of other tests in~\Cref{sec:other_tests}. The test statistic for this test is defined as: 

\begin{equation}
	W = \frac{(\sum_{i=1}^n a_i x_i)^2}{\sum_{i=1}^n (x_i - \overline{x})^2},
\end{equation}

where:
\begin{align*}
	x_i& \text{ is the $i$'th order statistic}, \\
	\bar{x}& \text{ is the sample mean}, \\
    m &= (m_1, \dots, m_n)^T \text{ are the expected values of}\\
            &\text{the order statistics},\\
    a &= \frac{m^TV^{-1}}{C}, \text{ for}\\
	V& \text{ the covariance matrix of }m,\\
    C &= || V^{-1} m ||.
	% (a_1, \dots, a_n) &= \frac{m^TV^{-1}}{C}, 
\end{align*}

The distribution of $W$ does not have a name, and the cutoff values for it are computed using Monte-Carlo simulations. $W$ lies between zero and one -- the closer it is to one, the more likely it is that the samples belong to the normal distribution. The $p$-value determines the confidence in the test statistic. 

%% file: sections/evaluation.tex
\section{Evaluation}

\subsection{Experimental Setting}

\textbf{Setting:} Certification is performed on a subsample of 500 images from the CIFAR 10 test set~\cite{krizhevsky_learning_2009}, each with 100 noise samples for selection ($n_0$), 10,000 noise samples for certification ($n$), and $\alpha = 0.001$. We use the same base model as Cohen et al.~\cite{cohen_certified_2019}, with a ResNet-110 architecture. \\

\noindent \textbf{Measurements:} Two types of results are collected: the first is the accuracy of the smoothed classifier, and the second is the size of the certification radius. The accuracy results attempt to determine if a naive defender will be able to detect the attack by simply observing the performance of the model. These are presented as a relative confusion matrix compared to the baseline unperturbed classifier, presenting the number of predictions that are correct, incorrect or abstains. The results for the impact of each attack on the certified radius were collected for the images where both the attacked model and the baseline predicted the same class. The fraction of images for which the relative certified radius ($R'/R$, where $R'$ is the manipulated radius and $R$ is the original radius) falls in each of the bins: (0, 1.0], (1, 1.1], (1.1, 1.25], (1.25, 1.5], (1.5, 2.0] and (2.0, $\infty$), is reported in~\Cref{tab:relative-radius}. The higher the value of $R'/R$, the easier it will be to find adversarial examples within the manipulated radius $R'$. According to Cohen et al.~\cite{cohen_certified_2019}, there is a 17\% probability of finding an adversarial example with $d_{max} = 1.5R$ and 53\% probability for $d_{max} = 2R$, where $d_{max}$ is as defined in~\Cref{eqn:evasion-attack}. \\

\noindent \textbf{Defences:} More sophisticated defences going beyond observing the change in performance of the classifier were also evaluated and are considered in~\Cref{sec:eval:defence}. First, the NIST test suite for certifying cryptographically secure \texttt{PRNG}s was run for various lengths of bit streams, ranging from $10^2$ to $10^6$, with 1000 tests in each run. The official version of the NIST test suite was used. For each \texttt{PRNG} tested, raw 64-bit integers are generated, and the binary representation is saved as ASCII-encoded strings in a text file. This is used as an input for the test utility. The next defence evaluated is the Shapiro-Wilk test for normality. First, raw 64-bit integers were transformed to the normal distribution using the default implementation in \texttt{numpy} \cite{harris_array_2020}, and then tested using the AS R94 algorithm \cite{royston_remark_1995} from the \texttt{scipy} library \cite{virtanen_scipy_2020}. This algorithm is limited to a maximum of 5000 samples, which is therefore the number of samples used to run our test. 

\subsection{Naive attacker}

The change in classifier accuracy with the naive attack is presented in~\Cref{tab:acc-dist}. As highlighted earlier, a smoothed model will never perform as well on the classification problem, as it becoming less accurate as more noise is introduced. Therefore, the attacker can get away with slightly reducing the accuracy and still escape detection since this is expected behaviour. $\mathcal{L}(0, 0.25)$ and $|\mathcal{N}(0, 0.25^2)|$ perform well as attack noise distributions in this regard, with only nominally reducing model accuracy. The other two attack distributions $\mathcal{U}(-0.25, 0.25)$ and $\mathcal{B}(0.5)$ result in significantly reduced accuracy, which will raise red flags for the defender.

\begin{table}[t]
	\centering
    	\begin{tabular}{cc|ccc}
    		 & \multicolumn{4}{c}{$\mathcal{L}(0, 0.25)$} \\
    		 \multirow{4}{*}{\rotatebox[origin=c]{90}{$\mathcal{N}(0, 0.25^2)$}} & & correct & incorrect & abstain \\
    		 \cline{2-5}
    		 & correct & \textbf{282} & 48 & 44 \\
    		 & incorrect & 9 & \textbf{52} & 19 \\
    		 & abstain & 6 & 20 & \textbf{20} \\
    	\end{tabular}
    	\caption*{Laplace distribution}
	\vspace{1em}
	\begin{tabular}{cc|ccc}
		& \multicolumn{4}{c}{$|\mathcal{N}(0, 0.25^2)|$} \\
		\multirow{4}{*}{\rotatebox[origin=c]{90}{$\mathcal{N}(0, 0.25^2)$}} & & correct & incorrect & abstain \\
		\cline{2-5}
		& correct & \textbf{287} & 73 & 14 \\
		& incorrect & 12 & \textbf{62} & 6 \\
		& abstain & 15 & 24 &\textbf{7} \\
	\end{tabular}
	\caption*{Absolute normal distribution}
	\vspace{1em}
	\begin{tabular}{cc|ccc}
		& \multicolumn{4}{c}{$\mathcal{U}(-0.25, 0.25)$} \\
		\multirow{4}{*}{\rotatebox[origin=c]{90}{$\mathcal{N}(0, 0.25^2)$}} & & correct & incorrect & abstain \\
		\cline{2-5}
		& correct &\textbf{128}& 235 & 11 \\
		& incorrect & 9 & \textbf{70} & 1 \\
		& abstain & 3 & 31 & \textbf{2} \\
	\end{tabular}
	\caption*{Uniform distribution}
	\vspace{1em}
	\begin{tabular}{cc|ccc}
		& \multicolumn{4}{c}{$\mathcal{B}(0.5)$} \\
		\multirow{4}{*}{\rotatebox[origin=c]{90}{$\mathcal{N}(0, 0.25^2)$}} & & correct & incorrect & abstain \\
		\cline{2-5}
		& correct &\textbf{154} & 113 & 107 \\
		& incorrect & 5 & \textbf{42} & 33 \\
		& abstain & 4 & 25 & \textbf{17}\\
	\end{tabular}
	\caption*{Bernoulli distribution}
    \caption{The above tables show a relative confusion matrix of the performance of \RS~subject to naive attacks, where the normal distribution is swapped for a different distribution. The uniform distribution leads to the greatest deviation from baseline performance, followed by the bernoulli distribution, then the absolute normal distribution, and finally the laplace distribution.}\label{tab:acc-dist}
\end{table}

Based on the discussion in~\Cref{sec:threat-model}, the probability of finding an adversarial example increases considerably as the relative manipulated radius increases. The focus here is on values of $R'/R$ that lie in (1.5, 2.0] and (2.0, $\infty$), which give the attacker a significantly high probability of performing an evasion attack. $|\mathcal{N}(0, 0.25^2)|$ performs really well in this regard, with 28\% of input images having a spoofed radius that falls in the highly vulnerable categories. $\mathcal{U}(-0.25,0.25)$ achieves the best attack success rate, but it is also the easiest attack distribution for the defender to detect by simply monitoring the performance of the classifier.

\begin{table*}[t]
	\centering
    \adjustbox{max width=\linewidth}{%
	\begin{tabular}{llcccccr}
        \toprule
		& $\frac{R'}{R} < 1$ & $1.0 < \frac{R'}{R} < 1.1$ & $1.1 < \frac{R'}{R} < 1.25$ & $1.25 < \frac{R'}{R} < 1.5$ & $1.5 < \frac{R'}{R} < 2.0$ & $\frac{R'}{R} > 2.0$ & max$\left( \frac{R'}{R} \right)$\\
        \midrule
        \multicolumn{8}{l}{\textit{Naive noise distribution replacement attack}}\\
		$\mathcal{L}(0, 0.25)$ & 0.80 & 0.07 & 0.02 & 0.02 & {\color{red} \textbf{0.02}} & {\color{red} \textbf{0.07}} & 6.03\\
		$|\mathcal{N}(0, 0.25^2)|$ & 0.24 & 0.31 & 0.09 & 0.08 & {\color{red} \textbf{0.10}} & {\color{red} \textbf{0.18}} & \textbf{76.25}\\
		$\mathcal{U}(-0.25, 0.25)$ & 0.32 & 0.22 & 0.03 & 0.08 & {\color{red} \textbf{0.07}} & {\color{red} \textbf{0.28}} & \textbf{13.99}\\
		$\mathcal{B}(0.5)$ & 0.88 & 0.02 & 0.01 & 0.02 & {\color{red} \textbf{0.03}} & {\color{red} \textbf{0.04}} & 9.58\\
		\midrule
        \multicolumn{7}{l}{\textit{Negative-Kurtosis attack}}\\
		$\alpha =  1$ & 0.58 & 0.31 & 0.05 & 0.03 & {\color{red} \textbf{0.02}} & {\color{red} \textbf{0.02}} & 2.10\\
		$\alpha =  2$ & 0.57 & 0.35 & 0.04 & 0.02 & {\color{red} \textbf{0.02}} & 0.00 & 1.38\\
		$\alpha =  3$ & 0.52 & 0.41 & 0.04 & 0.01 & {\color{red} \textbf{0.01}} & 0.00 & 1.06\\
		$\alpha =  4$ &0.55 & 0.40 & 0.03 & 0.01 & 0.00 & 0.00 & 1.95\\
        \midrule
        \multicolumn{8}{l}{\textit{Skewness attack}}\\
		$\beta =  0$ & 0.24 & 0.31 & 0.09 & 0.08 & {\color{red} \textbf{0.10}} & {\color{red} \textbf{0.18}} & \textbf{81.24}\\
		$\beta =  1$ & 0.24 & 0.35 & 0.10 & 0.08 & {\color{red} \textbf{0.09}} & {\color{red} \textbf{0.13}} & \textbf{36.27}\\
		$\beta =  2$ & 0.30 & 0.33 & 0.12 & 0.10 & {\color{red} \textbf{0.06}} & {\color{red} \textbf{0.08}} & \textbf{27.43}\\
		$\beta =  4$ & 0.34 & 0.39 & 0.12 & 0.06 & {\color{red} \textbf{0.04}} & {\color{red} \textbf{0.05}} & \textbf{18.11}\\
        \midrule
        \multicolumn{8}{l}{\textit{Positive-Kurtosis attack}}\\
		$\gamma =  0$ & 0.14 & 0.32 & 0.14 & 0.14 & {\color{red} \textbf{0.10}} & {\color{red} \textbf{0.16}} & \textbf{40.70}\\
		$\gamma = 1$ & 0.18 & 0.38 & 0.17 & 0.13 & {\color{red} \textbf{0.06}} & {\color{red} \textbf{0.08}} & \textbf{15.79}\\
		$\gamma = 2$  & 0.20 & 0.42 & 0.20 & 0.09 & {\color{red} \textbf{0.04}} & {\color{red} \textbf{0.06}} & \textbf{10.49}\\
		$\gamma = 4$  & 0.22 & 0.51 & 0.16 & 0.04 & {\color{red} \textbf{0.03}} & {\color{red} \textbf{0.04}} & 6.05\\
        \bottomrule
	\end{tabular}
    }
	\caption{This table shows the relative certified radius, $R'/R$, where $R'$ is the manipulated radius under attack, and $R$ is the baseline radius. The fraction of images for which $R'/R$ falls in different bins is shown for the naive attacks and all three of the bit-flipping \texttt{PRNG} attacks. The maximum value of the relative certified radius achieved for each attack is also reported. The values in red indicate instances when the attack managed to successfully manipulate the radius by a factor of at least 1.5. Out of the naive attacks, $\mathcal{U}(-0.25, 0.25)$ achieves the best performance, followed by $|\mathcal{N}(0, 0.25^2)|$, $\mathcal{L}(0, 0.25)$, and finally $\mathcal{B}(0.5)$. Among the bit-flipping \texttt{PRNG} attacks, the skewness performs the best, then the positive-kurtosis, followed by the negative-kurtosis.}\label{tab:relative-radius}
 % }
\end{table*}

\begin{table*}[!ht]
	\centering
	\begin{adjustbox}{width=   \linewidth}
		\begin{tabular}{lcccc|cccc|cccc|cccc}
            \toprule
			& \multirow{3}{*}{\rotatebox[origin=c]{270}{MT19937}} & \multirow{3}{*}{\rotatebox[origin=c]{270}{PCG64}} &              \multirow{3}{*}{\rotatebox[origin=c]{270}{Philox}} & \multirow{3}{*}{\rotatebox[origin=c]{270}{SFC64}} & & & & & & & & & & & & \\
                & & & & & \multicolumn{4}{c|}{$\alpha$} & \multicolumn{4}{c|}{$\beta$} & \multicolumn{4}{c}{$\gamma$} \\
			& & & & & 1 & 2 & 3 & 4 & 0 & 1 & 2 & 4 & 0 & 1 & 2 & 4 \\
			\midrule
			\textbf{Frequency} & 0 & 991 & 991 & 989 & 0 & 0 & 0 & 0 & 0 & 0 & 3 & \textbf{299} & 0 & 0 & 0 & 309 \\
                BlockFrequency & 0 & 991 & 989 & 989 & 2 & 0 & \textbf{667} & 0 & 915 & 990 & 993 & 996 & 922 & 986 & 995 & 990 \\
                \textbf{CumulativeSums} & 0 & 992 & 988 & 987 & 0 & 0 & 0 & 0 & 0 & 0 & 3 & \textbf{322} & 0 & 0 & 1 & \textbf{323} \\
                \textbf{Runs} & 0 & 990 & 988 & 991 & 0 & 0 & 0 & 0 & 0 & 0 & 102 & \textbf{805} & 0 & 0 & 0 & \textbf{233} \\
                LongestRun & 0 & 991 & 988 & 984 & 670 & 282 & \textbf{984} & 216 & 914 & 989 & 988 & 990 & 732 & 963 & 966 & 989 \\
                Rank & 0 & 991 & 990 & 992 & 995 & 994 & 992 & 985 & 993 & 983 & 992 & 986 & 988 & 986 & 993 & 994 \\
                FFT & 0 & 985 & 991 & 984 & 0 & 0 & 0 & 0 & 0 & 572 & 942 & 988 & 0 & 0 & 0 & \textbf{125} \\
                NonOverlappingTemplate & 0 & 980 & 982 & 981 & 0 & 0 & 0 & 0 & 116 & 825 & 929 & 975 & 0 & 8 & 302 & \textbf{787} \\
                OverlappingTemplate & 0 & 991 & 985 & 989 & 3 & 0 & 906 & 0 & 400 & 938 & 974 & 987 & 0 & 447 & 779 & 930 \\
                Universal & 0 & 996 & 988 & 984 & 921 & 150 & 978 & 0 & 987 & 989 & 982 & 993 & 980 & 989 & 986 & 990 \\
                ApproximateEntropy & 0 & 989 & 990 & 991 & 0 & 0 & 0 & 0 & 5 & 812 & 957 & 986 & 0 & 93 & 727 & 944 \\
                RandomExcursions & - & 618/628 & 613/626 & 617/631 & - & - & - & - & 2/2 & 24/26 & 63/64 & 210/215 & - & 26/27 & 83/85 & 220/224 \\
                RandomExcursionsVariant & - & 617/628 & 616/626 & 622/631 & - & - & - & - & 2/2 & 24/26 & 61/64 & 211/215 & - & 26/27 & 84/85 & 215/224 \\
                Serial & 0 & 988 & 987 & 977 & 0 & 0 & 196 & 0 & 901 & 985 & 990 & 990 & 297 & 945 & 984 & 982 \\
                LinearComplexity & 989 & 990 & 984 & 992 & 993 & 991 & 984 & 989 & 993 & 987 & 995 & 989 & 991 & 991 & 994 & 992 \\
            \bottomrule
		\end{tabular}
	\end{adjustbox}
	\caption{This table shows the results of the NIST test suite for cryptographically secure \texttt{PRNG}s on the three bit-flipping \texttt{PRNG} attacks with varying values of $\alpha$, $\beta$ and $\gamma$. The results for 4 popular modern \texttt{PRNG}s, MT19337, PCG64, Philox and SFC64 are also shown. PCG64 and SFC64 are NIST certified cryptographically secure \texttt{PRNG}s, MT19337 is the default \texttt{PRNG} in Python and Philox is the default \texttt{PRNG} in PyTorch. Tests were run for bit streams of length $10^6$ and the reported values are the number of instances of each test that passed out of 1000. The minimum pass rate for a good \texttt{RNG} recommended by NIST is 980. The frequency, cumulative sums and runs tests managed to detect all attacks with high confidence.}
 \label{tab:nist}
\end{table*}

\subsection{Bit-flipping \texttt{PRNG} attacker}

In this subsection we evaluate the three attacks described in \Cref{sec:method:PRNG}: negative-kurtosis, skewness and positive-kurtosis. The parameter values chosen for evaluation are $\alpha \in \left\{1, 2, 3, 4 \right\}$, $\beta \in \left\{0, 1, 2, 4 \right\}$ and $\gamma \in \left\{0, 1, 2, 4 \right\}$ for each of the attacks respectively. Note that none of these attacks can be detected by a naive defender. \Cref{apdx:sec:relative_acc} shows the relative accuracy results, that do not deviate significantly from the baseline. Relative certification radii for each attack are shown in~\Cref{tab:relative-radius}, with the skewness attack achieving the highest manipulated radii, followed by the positive-kurtosis. The negative-kurtosis attack performed worst, significantly altering the radius for only 4\% of images, even with $\alpha=1$.

\subsection{Defences}
\label{sec:eval:defence}

\textbf{NIST tests} -- \Cref{tab:nist} reports the number of bitstreams that passed each test out of 1000, where every bitstream was of length $10^6$. This excludes the random excursions and random excursions variant test, for which the NIST utility decides the number of bit streams to consider based on previous test results. The recommendation from NIST is that a good \texttt{PRNG} should pass at least 980 out of 1000 runs of each test. For the random excursion tests, the pass rate is reported based on the number of tests run. In addition to the attacked \texttt{PRNG}s, the results are also reported for the four most popular \texttt{PRNG}s currently in use: MT19937, PCG64, Philox and SFC64. While the other three manage to pass all tests, the MT19937 \texttt{PRNG} fails most of them, which is good news, since MT19937 is not recommended for any use case where security is required. 

\begin{figure*}[t]
    \centering
    \begin{subfigure}{0.33\linewidth}
        \includegraphics[width=\linewidth]{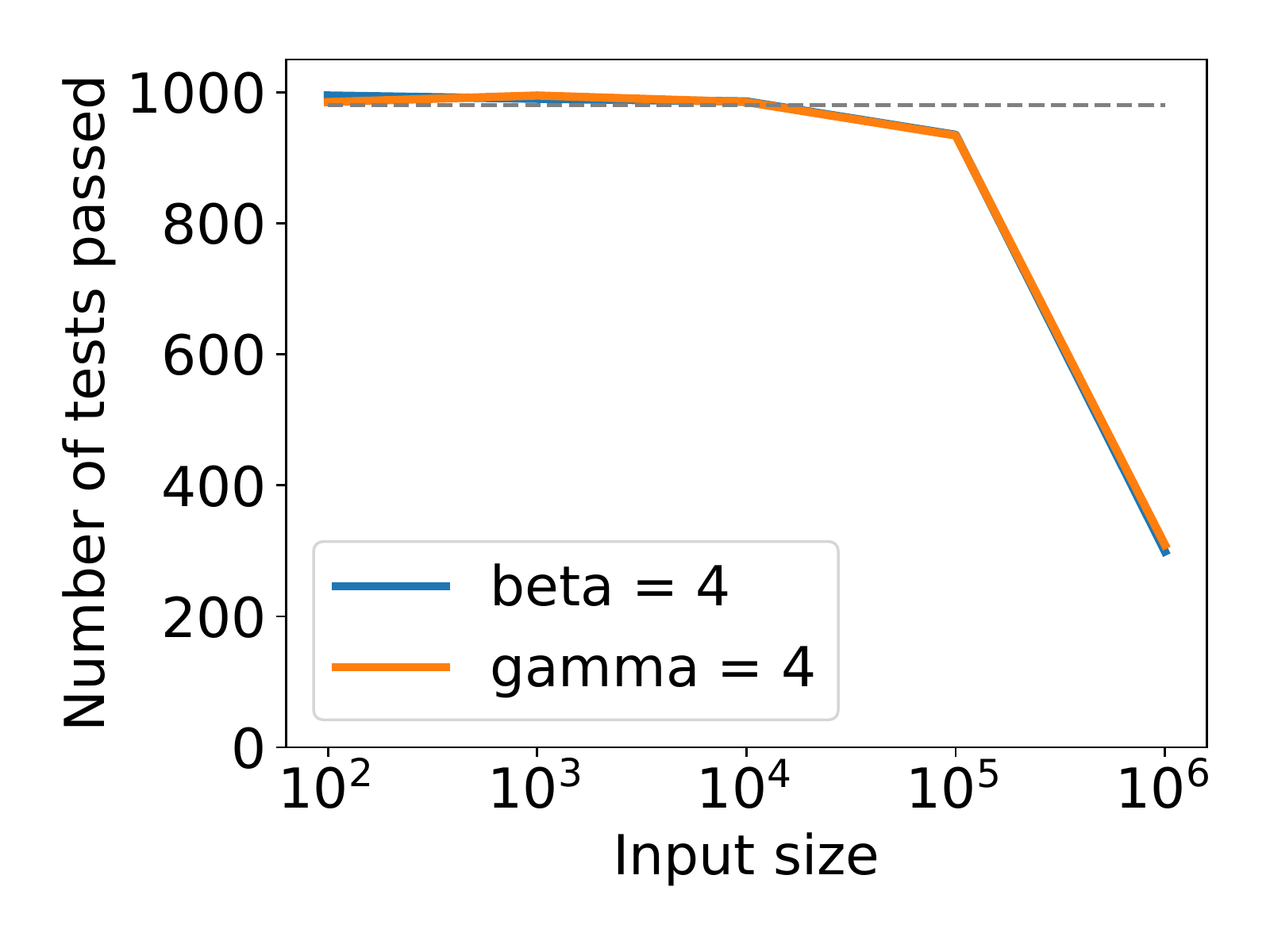}
        \caption{Frequency}
        \label{fig:frequency}
    \end{subfigure}
    \begin{subfigure}{0.33\linewidth}
        \includegraphics[width=\linewidth]{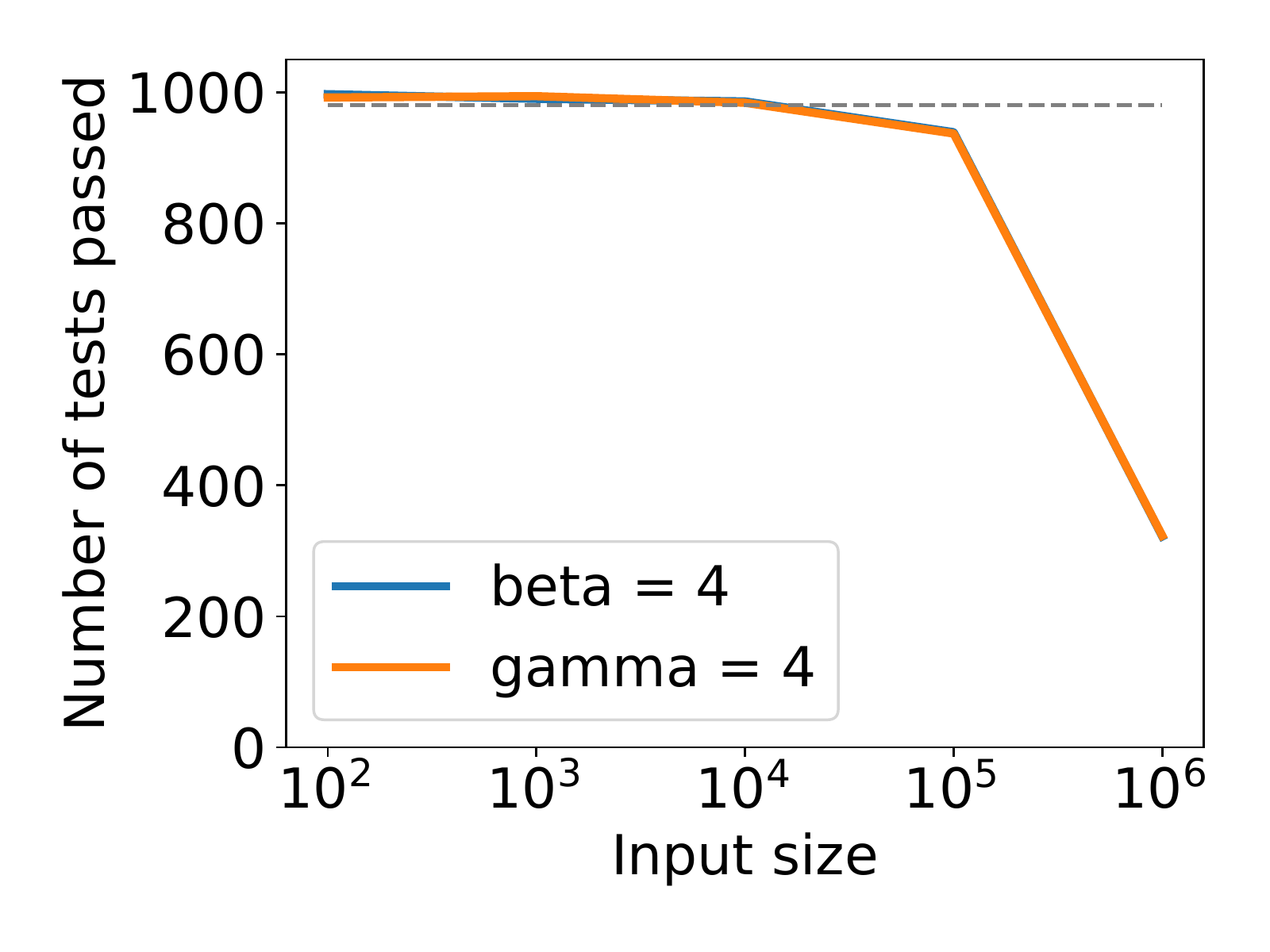}
        \caption{Cumulative Sums}
        \label{fig:cumsum}
    \end{subfigure}
    \begin{subfigure}{0.33\linewidth}
        \includegraphics[width=\linewidth]{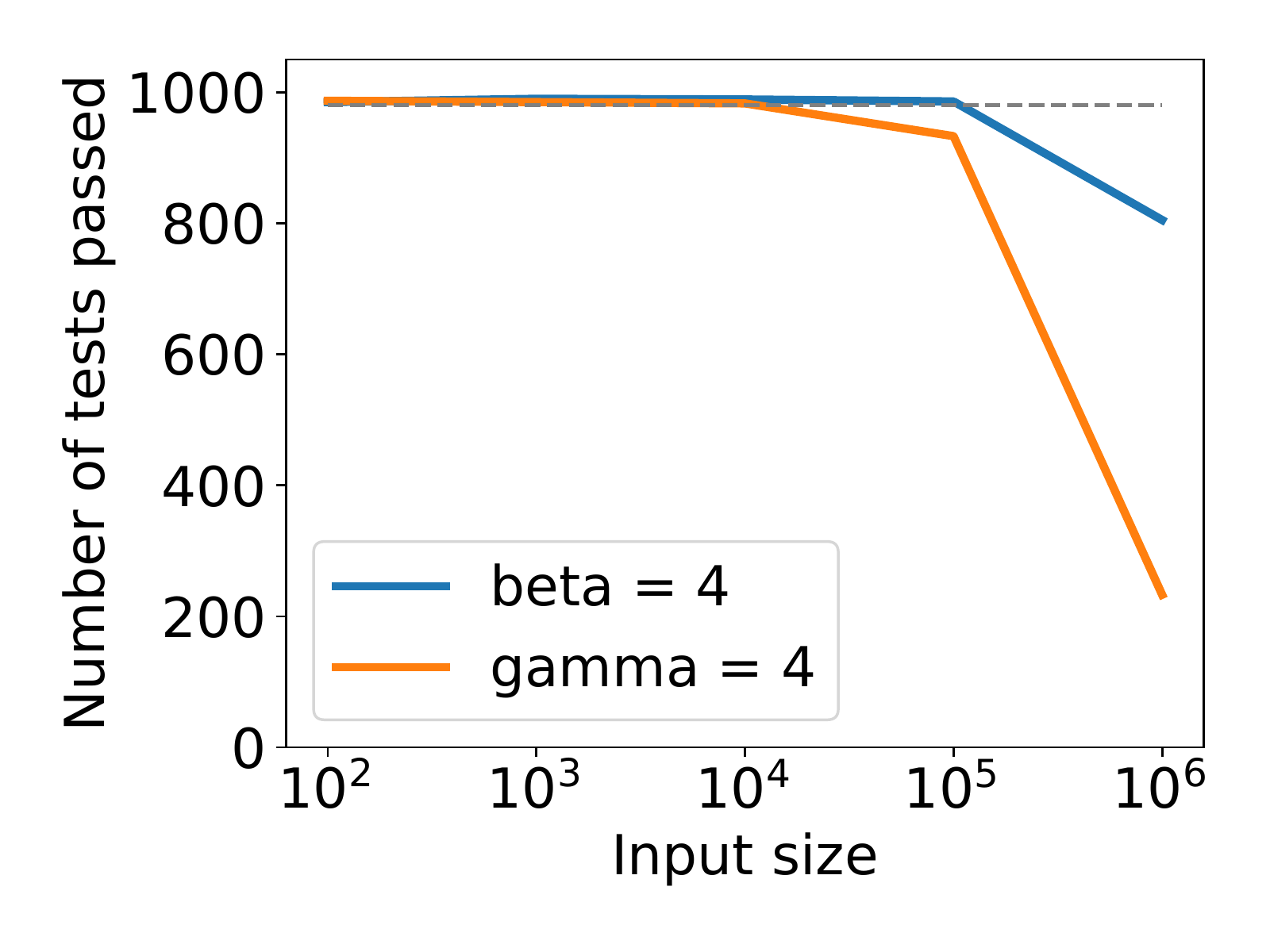}
        \caption{Runs}
        \label{fig:runs}
    \end{subfigure}
    \caption{These figures show the NIST test pass rates for the frequency, cumulative sums and runs tests, the three tests that managed to detect all the attacked \texttt{PRNG}s with high confidence at an input size of $10^6$. These plots show how the pass rates vary as the input size is increased from the minimum recommended value by NIST, $10^2$, to $10^6$ for the two attacks that were the hardest to detect: the skewness attack with $\beta=4$ and the positive-kurtosis attack with $\gamma=4$. The attacks become detectable only after the input size is increased to $> 10^5$.}
    \label{fig:nist_vs_input}
\end{figure*}

Focusing on the results for $\beta=4$ and $\gamma=4$, the tests that were able to successfully detect both the skewness and positive-kurtosis attacks with high confidence are the frequency test, cumulative sums test and the runs test. It must be noted that the minimum recommended input size by NIST for these three tests is 100, which is significantly lower than the input size of $10^6$ used for generating the results in \Cref{tab:nist}. \Cref{fig:nist_vs_input} shows the pass rate for these tests for different sample sizes ranging from $10^2$ to $10^6$ for the attacked \texttt{PRNG}s with $\beta=4$ and $\gamma=4$. Results show a significant drop in pass rate only after the input size is increased above $10^5$, demonstrating a clear need to re-evaluate the recommended parameters and input sizes suggested by NIST for safety-critical applications.

Another important result is that for the negative-kurtosis attack with $\alpha=3$. It was able to get a significantly higher pass rate compared to the rest of the \texttt{PRNG}s for the same attack. It goes to show that statistical tests cannot always be relied upon and bad \texttt{PRNG}s sometimes get a high pass rates. \\

\begin{figure*}
\begin{subfigure}{0.33\linewidth}
	\centering
	\begin{subfigure}{0.45\linewidth}
		\includegraphics[width=0.95\linewidth]{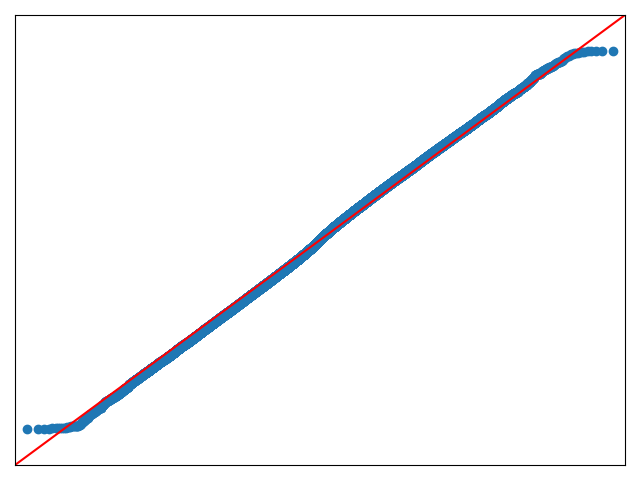}
		\caption*{$\alpha = 1$}
	\end{subfigure}
	\begin{subfigure}{0.45\linewidth}
		\includegraphics[width=0.95\linewidth]{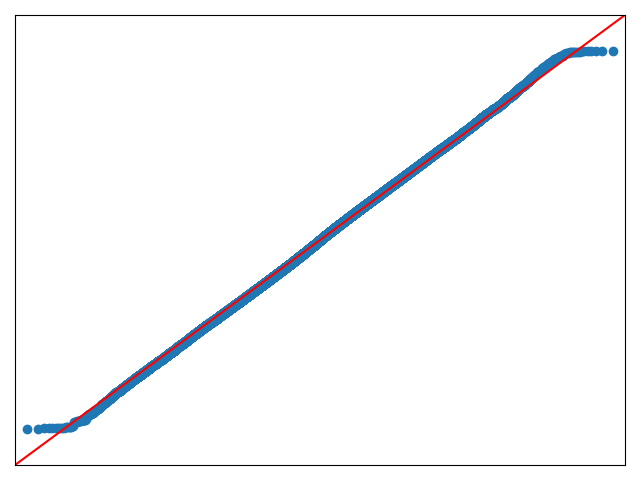}
		\caption*{$\alpha = 2$}
	\end{subfigure}
	\begin{subfigure}{0.45\linewidth}
		\includegraphics[width=0.95\linewidth]{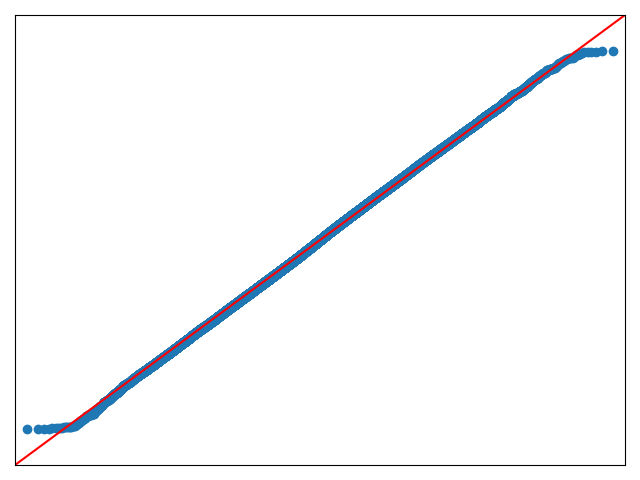}
		\caption*{$\alpha = 3$}
	\end{subfigure}
	\begin{subfigure}{0.45\linewidth}
		\includegraphics[width=0.95\linewidth]{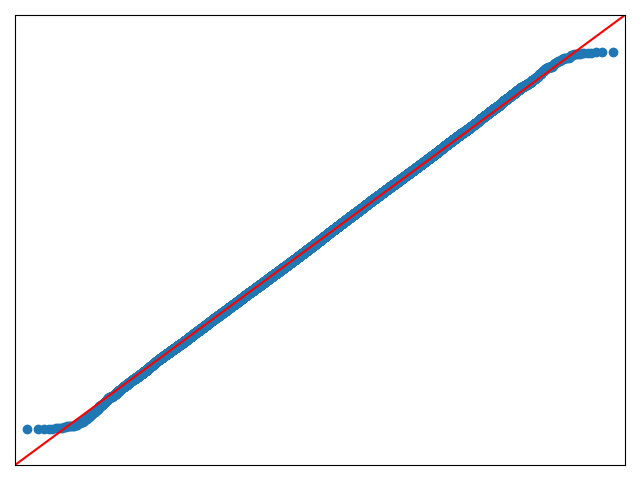}
		\caption*{$\alpha = 4$}
	\end{subfigure}
	\caption{QQ plots for Negative-Kurtosis Attack}
	\label{fig:qq-skewness}
\end{subfigure}\quad
\begin{subfigure}{0.33\linewidth}
	\centering
	\begin{subfigure}{0.45\linewidth}
		\includegraphics[width=0.95\linewidth]{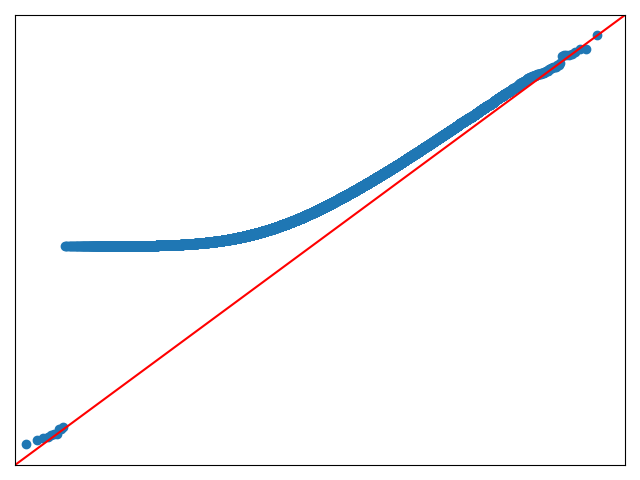}
		\caption*{$\beta = 0$}
	\end{subfigure}
	\begin{subfigure}{0.45\linewidth}
		\includegraphics[width=0.95\linewidth]{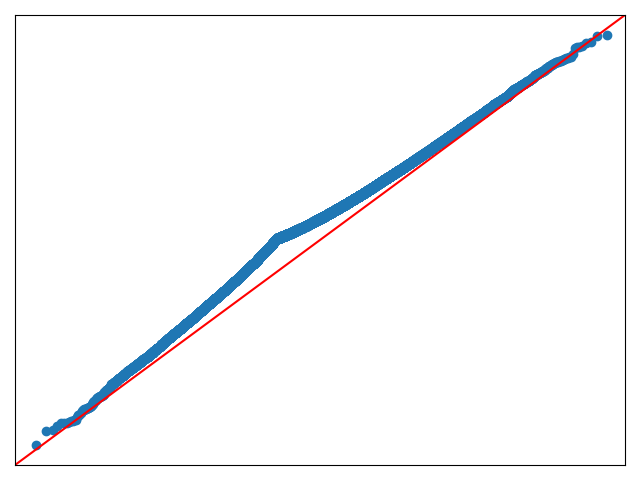}
		\caption*{$\beta = 1$}
	\end{subfigure}
	\begin{subfigure}{0.45\linewidth}
		\includegraphics[width=0.95\linewidth]{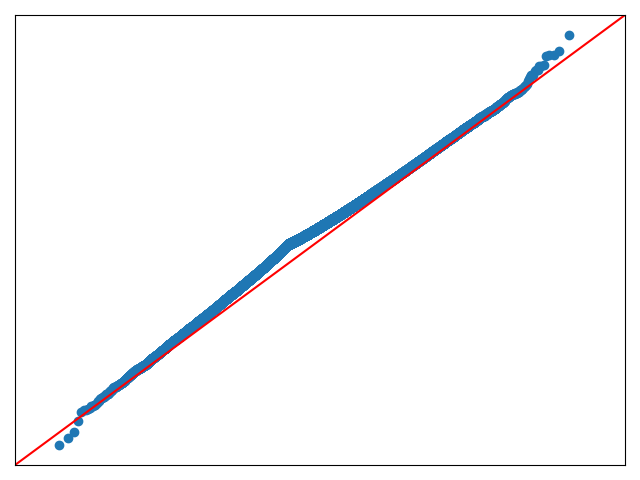}
		\caption*{$\beta = 2$}
	\end{subfigure}
	\begin{subfigure}{0.45\linewidth}
		\includegraphics[width=0.95\linewidth]{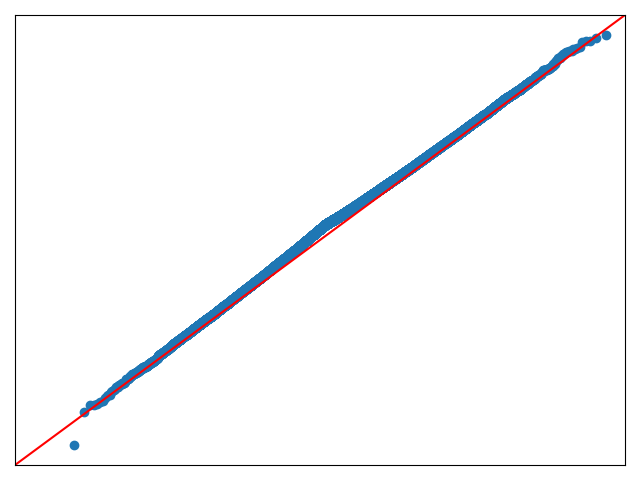}
		\caption*{$\beta = 4$}
	\end{subfigure}
	\caption{QQ plots for Skewness Attack}
	\label{fig:qq-skewness}
\end{subfigure}\quad
\begin{subfigure}{0.33\linewidth}
	\centering
	\begin{subfigure}{0.45\linewidth}
		\includegraphics[width=0.95\linewidth]{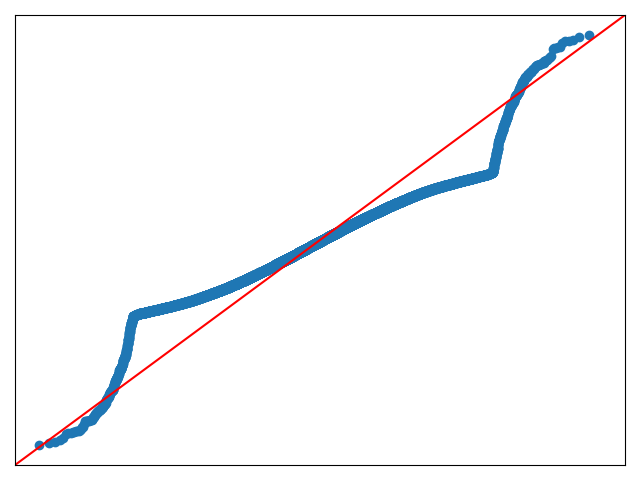}
		\caption*{$\gamma = 0$}
	\end{subfigure}
	\begin{subfigure}{0.45\linewidth}
		\includegraphics[width=0.95\linewidth]{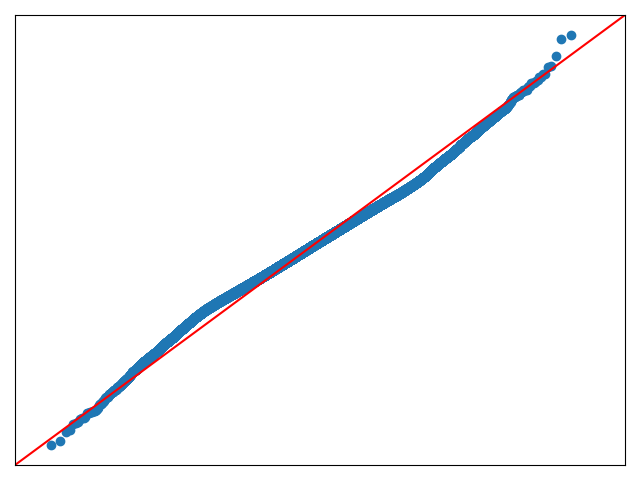}
		\caption*{$\gamma = 1$}
	\end{subfigure}
	\begin{subfigure}{0.45\linewidth}
		\includegraphics[width=0.95\linewidth]{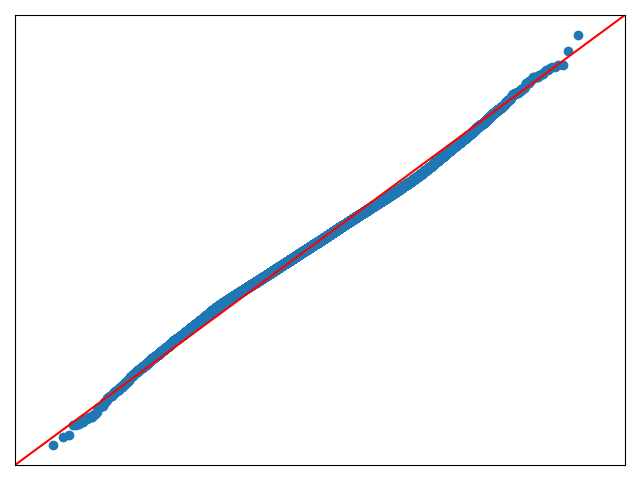}
		\caption*{$\gamma = 2$}
	\end{subfigure}
	\begin{subfigure}{0.45\linewidth}
		\includegraphics[width=0.95\linewidth]{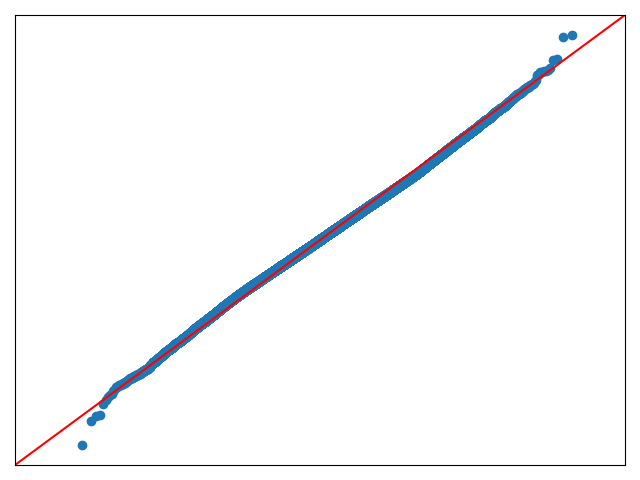}
		\caption*{$\gamma = 4$}
	\end{subfigure}
	\caption{QQ plots for Positive-Kurtosis Attack}
	\label{fig:qq-kurtosis}
\end{subfigure}
\caption{These figures show the quantile-quantile (QQ) plots for $10^5$ samples collected from \texttt{PRNG}s subject to the negative-kurtosis, skewness and positive-kurtosis attacks, showing the difference in CDF relative to the normal distribution. While attacks that significantly alter the distribution can be detected visually, this is not always apparent, as is the case with the skewness attack with $\beta=4$ and positive-kurtosis attacks with $\gamma=2$ and $\gamma=4$.}
\label{fig:qq_plots}
\end{figure*}

\noindent \textbf{Shapiro test} The results from the NIST tests demonstrate that it is difficult to detect attacks against \texttt{PRNG}s that alter the distribution once it is transformed to the normal distribution by using existing tests developed for ensuring that \texttt{PRNG}s are cryptographically secure. While attacks such as the negative kurtosis attack that significantly alter the distribution of raw random integers produced by the \texttt{PRNG} can be detected, more sophisticated attacks such as the skewness attack and the negative kurtosis attack escape detection since they modify bits that have a greater impact on the normal distribution transformation. Therefore, the next class of defences explored here focuses on detecting the attack post-transformation, and is only performed for the skewness and positive kurtosis attacks.

The first step in verifying whether a sampled distribution is close to normal is to manually look at the quantile-quantile (Q-Q) plot of the samples. It shows the difference between the sample cumulative distribution and the expected CDF. A QQ plot of the normal distribution sampled from the baseline PCG \texttt{PRNG} is shown in~\Cref{fig:qq-baseline} in the Appendix. The QQ plots of \texttt{PRNG}s attacked with the negative kurtosis, skewness and positive kurtosis attacks are shown in~\Cref{fig:qq_plots}. It is clear from the plots that with lower values of $\beta$ and $\gamma$, it is fairly easy to identify that the \texttt{PRNG} is not behaving as expected. However, with higher values of $\beta$ and $\gamma$, which do not alter the distribution as much, it is a non-trivial task.

Shapiro-Wilk test results are reported in \Cref{tab:shapiro}. While the rest of the \texttt{PRNG}s have extremely low $p$-values and therefore can be considered to have significantly deviated from the normal distribution, the \texttt{PRNG} subject to the positive kurtosis attack achieved a $W$ value that is the same as the baseline within three significant digits and also manages to get a $p$-value of 0.045 -- \ie~it passes the NIST recommended confidence threshold. Just as the NIST test suite was not a foolproof method of detecting deviations from the uniform distribution, the Shapiro-Wilk test, while useful, is not a foolproof technique for detecting deviation from the normal distribution.

\begin{table}[t]
	\centering
	\begin{tabular}{lcr}
		& Test Statistic & $p$-value\\
		\toprule
		PCG Baseline & 0.999 & 0.203 \\
		% \midrule
  %           $\alpha = 1$ & 0.955 & $7.78 \times 10^{-37}$ \\
  %           $\alpha = 2$ & 0.952 & $9.07 \times 10^{-38}$ \\
  %           $\alpha = 3$ & 0.955 & $1.18 \times 10^{-36}$ \\
  %           $\alpha = 4$ & 0.056 & $2.23 \times 10^{-36}$ \\
            \midrule
		$\beta = 0$ & 0.925 & $2.66\times10^{-44}$ \\
		$\beta = 1$ & 0.988 & $5.71\times10^{-20}$ \\
		$\beta = 2$ & 0.992 & $1.09\times10^{-15}$ \\
		$\beta = 4 $ & 0.997 & $2.34\times10^{-6}$ \\
		\midrule
		$\gamma = 0$ & 0.983 & $4.93\times10^{-24}$ \\
		$\gamma = 1$ & 0.996 & $1.04\times10^{-9}$ \\
		$\gamma = 2$ & 0.998 & $3.17\times10^{-5}$ \\
		$\mathbf{\gamma = 4}$ &\textbf{0.999} & $\mathbf{0.045}$ \\
        \bottomrule
	\end{tabular}
	\caption{This table shows the value of the Shapiro-Wilk test statistic and p-value for tests conducted on 5000 samples collected from the baseline PCG \texttt{PRNG}, and \texttt{PRNG}s subject to the skewness and positive-kurtosis attacks that were difficult to detect with the NIST test suite. While other attacks could be detected with the Shapiro-Wilk test, the positive-kurtosis attack with $\gamma = 4$ managed to achieve a $p$-value that is greater than 0.01 -- the threshold recommended by NIST for its tests.}
	\label{tab:shapiro}
\end{table}

%% file: sections/discussion.tex
% Explore the broader impact and contribution of this work in the discussion section;
% Tone Down Claims Regarding New Standard
% Practical Implementation of Attack

\section{Discussion}

This work successfully demonstrates the feasibility of an attack on the pseudo-random number generator in an ML library to manipulate the certified radius obtained for the industry standard~\RS~technique. By altering the bit stream produced by the random number generator, kurtosis or skewness can be introduced into the noise distribution used. This can significantly affect the certified radius and thus make the end user over-estimate the real robustness. This work adds to a small but growing area of research into machine learning security that looks beyond the attack surfaces of the models themselves and considers systems that enable them. 

\noindent\textbf{Broader Impact} ML security is just like traditional security, in that practitioners need to consider the whole technical stack and not just limit themselves to the `ML' parts of data collection, training and inference-time attacks. An ML model runs on the same kind of hardware and software layers as every other piece of software and is therefore just as vulnerable to these `traditional' attack surfaces. The literature already notes that attacks can come from ML compilers~\cite{clifford_impnet_2022}, underlying platforms~\cite{rakin2019bitflip,boutros2020neighbors}, model architectures~\cite{bober-irizar_architectural_2022}, or even quantisation granularity~\cite{ma2023quantization}. Yet there is no standard guide for secure ML deployment, making responsible deployment challenging. 

\noindent \textbf{Protocols and guidelines} New protocols and guidelines must be designed specifically for current machine learning practices. These should recognise that firms outsource parts of the ML pipeline during design, training and inference. Current industry protocols trust the software and hardware on which these models run, without a clear threat model. Our paper helps show that this leads to exploitable vulnerabilities. 

\noindent \textbf{Randomness in ML} The meaning of a \textit{secure} \texttt{PRNG} needs to be updated for ML models. As these become more complex and place more reliance on non-deterministic algorithms, the random number generator is becoming as important in these systems as it is in cryptography. Some of the generators commonly used in ML are derived from those used in cryptography, but have been weakened to make them run faster. But even a cryptographically secure generator is not necessarily secure for ML, as the output is often shaped to have a distribution other than the uniform one, and this shaping provides a new attack vector. So statistical tests for the actual distributions commonly used in machine learning should be explicitly incorporated into official benchmarks.\\

\noindent \textbf{Limitations and Realism} Although access to randomness generators is a strong adversary assumption, it is not unrealistic. The dual EC generator was backdoored by NSA in the past to break cryptography that used it. In this paper we demonstrate that randomness can similarly be used to break ML applications and it should be included in threat models.\\

\noindent \textbf{Practical Attacks} In this paper we launch our attacks using the random number generator -- in practice they can also be utilised at different points in the technical stack. For example, an adversary can use BitFlips to change random numbers passing over the PCI-e bus \cite{rakin2019bitflip}; if randomness comes from the host platform, they could use VPGA attacks~\cite{boutros2020neighbors} or Rowhammer~\cite{kim2014rowhammer} to achieve the same effect. \\

\noindent \textbf{On passing tests} Bassham et al. stress \cite{bassham_iii_sp_2010} that by the very nature of statistical tests, some good \texttt{PRNG}s will fail them, while some bad \texttt{PRNG}s will pass. For example, we find that cryptographically secure PCG64 fails Lilliefors test at $10^5$ samples (\Cref{tab:norm100}). Therefore, the design of a new \texttt{PRNG} should be mathematically and architecturally sound, allow for external scrutiny, and its implementation must be studied carefully for backdoors.  Other applications of randomness in ML are also likely to have specific vulnerabilities and will need tests designed to target them~\eg~in differential privacy.\\

\noindent \textbf{Better Standards} We argue that ML needs better standards for randomness. This must be informed by a better threat model, as well as application domain knowledge. While a cryptographically secure \texttt{PRNG} can sometimes meet this requirement, ML developers often look for better performance, and their requirements for randomness differ in other ways from those of cryptographers. ML requires that the sampling distribution is accurate whereas cryptographers aim to reduce predictability. So while there may be an overlap between the requirements for the two applications, there is still a need for ML-specific randomness guidance. For example, if a given ML application uses the \texttt{PRNG} to sample Gaussian noise, then this property must be explicitly verified. \Cref{sec:other_tests} shows that each normality test detects different attacks, while no test detects all of them reliably. A comprehensive test suite should become part of the standard with a well-calibrated sample size \eg~$10^5$ for Gaussian noise-dependent ML workloads and multiple Guassian noise-specific tests.

ML brings up novel issues that were never a concern of cryptographers. In particular, we need to revisit issues around floating-point representations of random numbers -- Mironov noted these in 2012~\cite{mironov2012significance}, and Zhuang et al. found that similar attacks work in 2022~\cite{zhuang2021randomness}. Our attacks further demonstrate the importance of these issues. In cryptographic applications, key generation is performed relatively rarely, so an application developer who is suspicious of the quality of random numbers supplied by the platform will pass them through a \texttt{PRNG} of their own construction, which typically maintains a pool of randomness and used one-way functions both to update this pool and to draw key material from it. However, this uses several Kb of memory and several thousand instructions for every pseudorandom value drawn. Many ML applications cannot afford to do this as they make intensive use of computation, draw many random values, and can waste neither compute cycles nor GPU-adjacent RAM. This rules out the use of application-specific \texttt{PRNG}s, and also rules out continuous testing of the random numbers supplied by the platform -- running the NIST suite already takes hours. 

As such testing is impractical at runtime we may well therefore need separate test suites for system certification and to detect runtime attacks, as in some cryptographic systems; but that may not be enough. In critical applications, designers may need to understand that mechanisms for shaping bitstream statistics are within the trusted computing base. System evaluation will also have to take account of this and demand assurance against the kind of attacks discussed here.

%% file: sections/conclusion.tex
\section{Conclusion}

Machine learning systems rely on randomness for many purposes, ranging from differential privacy, through selecting representative subsets of data for training, to privacy amplification in federated learning and the generation of synthetic data. Yet the consequences of poor random number generators have remained unstudied. Common tools optimise random number generation for speed rather than security, and no attention is paid to the possibility that adversaries might manipulate random number sources to undermine model safety or security. In this paper we presented two proof-of-concept attacks on randomised smoothing, showing first that replacing a Gaussian random generator with a Laplacian one is effective if detectable, and second that flipping targeted bits in a Gaussian random generator is also effective but much more covert, in the sense of being very hard to detect using standard NIST randomness tests with default parameters.

The consequences, and future directions of research, are three-fold. First of all, there is a real need to define what security means for an ML \texttt{PRNG}, just as there has been extensive study of the requirements for cryptographic \texttt{PRNG}s. Second, we need to put more effort into exploring attacks that modify the random bit-stream in less predictable ways to improve attack performance and reduce the probability of detection. Finally, we need proper standards for randomness in ML systems and indeed in statistics generally -- for which a careful study of both attack and defence is the necessary foundation.

%% file: sections/appendix.tex
\appendix
\section*{Appendix}
\label{sec:appendix}

\section{Implementation of Bit-flipping \texttt{PRNG} attacks}

\begin{lstlisting}[language=C, caption={Original function to generate random integer}, label=lst:pcg-original]
static inline uint64_t pcg64_next64(pcg64_state *state) {
  return pcg64_random_r(state->pcg_state);
}
\end{lstlisting}

\begin{lstlisting}[language=C, caption={Negative-Kurtosis attack}, label=lst:pcg-kurtosis]
static inline uint64_t pcg64_next64(pcg64_state *state) {
  uint64_t rnd = pcg64_random_r(state->pcg_state);
  // convert to float
  double double_rnd = (rnd >> 11) * (1.0 / 9007199254740992.0);
  double a = 1.0 / state->alpha;
  double b = 1.0 - a / 2.0;
  // skew uniform distribution
  double_rnd = (sqrt(b * b + 2 * double_rnd * a) - b) / a;
  // convert back to int
  uint64_t new_rnd = (uint64_t) (double_rnd * 9007199254740992.0);
  // shift left and replace sign and idx bits with original random bits
  rnd = (new_rnd << 9) + (rnd & 0x1ff);
  return rnd;
}
\end{lstlisting}

\begin{lstlisting}[language=C, caption={Skewness attack}, label=lst:pcg-skew]
static inline uint64_t pcg64_next64(pcg64_state *state) {
    uint64_t rnd = pcg64_random_r(state->pcg_state);
    // check if the sign bit is set in random number and counter is equal to beta
    if ((rnd >> 8) & 0x1) {
        if (state->skewness >> state->beta) {
            // reset counter
            state->skewness = 0x1;
            // set sign bit to zero
            rnd &= ~(0x1 << 8)
        } else {
            // increment counter
            state->skewness <<= 1;
        }
    }
    return rnd;
}
\end{lstlisting}

\begin{lstlisting}[language=C, caption={Positive-Kurtosis attack}, label=lst:pcg-zero]
	static inline uint64_t pcg64_next64(pcg64_state *state) {
		uint64_t rnd = pcg64_random_r(state->pcg_state);
		// check if counter is equal to gamma
		if (state->kurtosis >> state->gamma) { 
			// reset counter 
			state->zero = 0x1;
			// right shift idx bits
			rnd = (rnd && ~0xff) & ((rnd & 0xff) >> 1);
		} else {
			// increment counter
			state->kurtosis <<= 1;
		}
		return rnd;
	}
\end{lstlisting}

\section{Relative Accuracy for Bit-flipping \texttt{PRNG} Attacks}
\label{apdx:sec:relative_acc}

\begin{table}[H]
    \small
	\centering
        \scriptsize
	\caption{Relative accuracy of skewness attack}\label{tab:skew-acc}
	\subcaption{$\beta = 0$}
	\begin{tabular}{c|ccc}
		& correct & incorrect & abstain \\
		\hline
		correct & \textbf{287} & 73 & 14 \\
		incorrect & 12 & \textbf{62} & 6 \\
		abstain & 15 & 24 & \textbf{7} \\
	\end{tabular}
	\subcaption{$\beta = 1$}
	\begin{tabular}{c|ccc}
		& correct & incorrect & abstain \\
		\hline
		correct & \textbf{341} & 23 & 10 \\
		incorrect & 6 & \textbf{64} & 10 \\
		abstain & 11 & 22 & \textbf{13} \\
	\end{tabular}
	\subcaption{$\beta = 2$}
	\begin{tabular}{c|ccc}
		& correct & incorrect & abstain \\
		\hline
		correct & \textbf{354} & 13 & 7 \\
		incorrect & 2 & \textbf{73} & 5 \\
		abstain & 10 & 18 & \textbf{18} \\
	\end{tabular}
	\subcaption{$\beta = 4$}
	\begin{tabular}{c|ccc}
		& correct & incorrect & abstain \\
		\hline
		correct & \textbf{363} & 4 & 7 \\
		incorrect & 0 & \textbf{74} & 6 \\
		abstain & 8 & 13 & \textbf{25} \\
	\end{tabular}
\end{table}

\begin{table}[H]
	\centering
        \scriptsize
	\caption{Relative accuracy for the positive kurtosis attack}\label{tab:pos-kurt-acc}
	\subcaption{$\gamma = 0$}
	\begin{tabular}{c|ccc}
		& correct & incorrect & abstain \\
		\hline
		correct & \textbf{349} & 19 & 6 \\
		incorrect & 6 & \textbf{67} & 7 \\
		abstain & 15 & 21 & \textbf{10} \\
	\end{tabular}
	\subcaption{$\gamma = 1$}
	\begin{tabular}{c|ccc}
		& correct & incorrect & abstain \\
		\hline
		correct & \textbf{346 }& 21 & 7 \\
		incorrect & 6 & \textbf{68} & 6 \\
		abstain & 15 & 20 & \textbf{11} \\
	\end{tabular}
	\subcaption{$\gamma = 2$}
	\begin{tabular}{c|ccc}
		& correct & incorrect & abstain \\
		\hline
		correct & \textbf{362} & 3 & 9 \\
		incorrect & 2 & \textbf{68} & 10 \\
		abstain & 11 & 16 & \textbf{19} \\
	\end{tabular}
	\subcaption{$\gamma = 4$}
	\begin{tabular}{c|ccc}
		& correct & incorrect & abstain \\
		\hline
		correct & \textbf{369} & 1 & 4 \\
		incorrect & 0 & \textbf{75} & 5 \\
		abstain & 6 & 7 & \textbf{33} \\
	\end{tabular}
\end{table}

\begin{table*}[t]
\centering
\adjustbox{max width=\linewidth}{
\begin{tabular}{lcccccc}
% \hline
\toprule
 & D'Agostino & Kolmogorov-Smirnov & Cramer-von Mises & Jarque-Bera & Shapiro & Lilliefors \\
% \midrule
% \toprule
PCG64 & - & - & - & - & - & \textbf{$10^5$} \\
\midrule
$\alpha = 1$ & $10^3$ & $10^2$ & $10^3$ & $10^3$ & $10^3$ & $10^3$ \\
$\alpha = 2$ & $10^4$ & $10^3$ & $10^3$ & $10^4$ & $10^4$ & $10^4$ \\
$\alpha = 3$ & $10^4$ & $10^3$ & $10^4$ & $10^4$ & $10^4$ & $10^4$ \\
$\alpha = 4$ & $10^4$ & $10^4$ & $10^4$ & $10^4$ & $10^4$ & $10^4$ \\
\midrule
$\beta = 0$ & $10^2$ & $10^2$ & $10^2$ & $10^2$ & $10^2$ & $10^2$ \\
$\beta = 1$ & $10^2$ & $10^2$ & $10^2$ & $10^2$ & $10^2$ & $10^2$ \\
$\beta = 2$ & $10^2$ & $10^2$ & $10^2$ & $10^2$ & $10^2$ & $10^2$ \\
$\beta = 4$ & $10^2$ & $10^2$ & $10^2$ & $10^2$ & $10^2$ & $10^2$ \\
\midrule
$\gamma = 0$ & $10^2$ & $10^2$ & $10^2$ & $10^2$ & $10^2$ & $10^4$ \\
$\gamma = 1$ & $10^2$ & $10^2$ & $10^3$ & $10^2$ & $10^2$ & $10^4$ \\
$\gamma = 2$ & $10^2$ & $10^3$ & $10^3$ & $10^2$ & $10^2$ & $10^4$ \\
$\gamma = 4$ & $10^2$ & $10^3$ & $10^3$ & $10^2$ & $10^2$ & $10^4$ \\
\bottomrule
\end{tabular}
}
\caption{This table reports the minimum sample size for which tests started to fail for each \texttt{PRNG} compared to PCG64. 1000 tests were run for increasing sample sizes from $10^2$ to $10^6$ following the same pass criteria as the NIST test suite in~\Cref{tab:nist}. A sample size of $\mathbf{10^5}$ was required to see every attacked \texttt{PRNG} conclusively fail all 6 tests.}
\label{tab:norm100}
\end{table*}

\begin{table}[H]
	\centering
        \scriptsize
	\caption{Relative accuracy for negative kurtosis attack}\label{tab:neg-kurt-acc}
	\subcaption{$\alpha = 1$}
	\begin{tabular}{c|ccc}
		& correct & incorrect & abstain \\
		\hline
		correct & \textbf{361} & 0 & 13 \\
		incorrect & 2 & \textbf{69} & 9 \\
		abstain & 6 & 4 & \textbf{36} \\
	\end{tabular}
	\subcaption{$\alpha = 2$}
	\begin{tabular}{c|ccc}
		& correct & incorrect & abstain \\
		\hline
		correct & \textbf{369} & 0 & 5 \\
		incorrect & 0 & \textbf{73} & 7 \\
		abstain & 3 & 2 & \textbf{41} \\
	\end{tabular}
	\centering
	\subcaption{$\alpha = 3$}
	\begin{tabular}{c|ccc}
		& correct & incorrect & abstain \\
		\hline
		correct & \textbf{371} & 0 & 3 \\
		incorrect & 0 & \textbf{75} & 5 \\
		abstain & 3 & 1 & \textbf{42} \\
	\end{tabular}
	\subcaption{$\alpha = 4$}
	\begin{tabular}{c|ccc}
		& correct & incorrect & abstain \\
		\hline
		correct & \textbf{370} & 0 & 4 \\
		incorrect & 0 & \textbf{77} & 3 \\
		abstain & 1 & 1 & \textbf{44} \\
	\end{tabular}
\end{table}

\section{Baseline QQ plot for PCG64 \texttt{PRNG}}

\begin{figure}[!ht]
	\centering
	\includegraphics[width=0.5\linewidth]{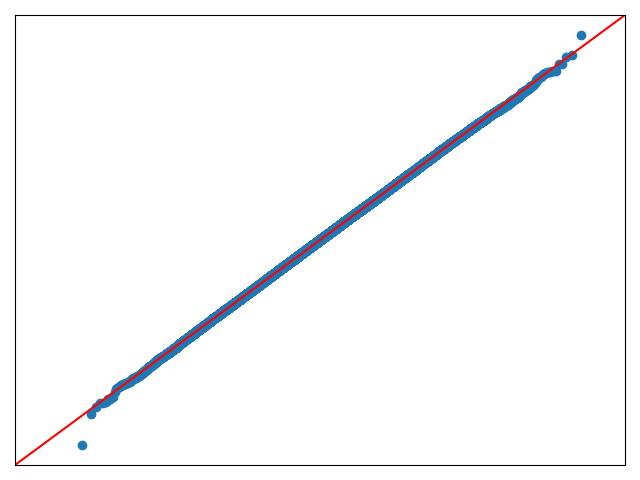}
	\caption{Baseline QQ plot}
	\label{fig:qq-baseline}
\end{figure}

\section{Other normality tests}
\label{sec:other_tests}

\Cref{tab:norm100} show the results of the following tests for normality with different sample sizes on the \texttt{PRNG} attacks: D'Agostino's K-squared test, Kolmogorov-Smirnov test, Cramer-von Mises criterion, Jarque-Bera test, Shapiro-Wilk test, and Lilliofors test. 1000 instances of each test were run for varying sample sizes from $10^2$ to $10^6$. A test was passed if the $p$-value reported was greater than 0.01 and the overall pass threshold was set as 980/1000 tests according to NIST's recommendations. \Cref{tab:norm100} reports the minimum sample size for which each test failed for the different attacked \texttt{PRNG}s. Test results for a baseline PCG64 \texttt{PRNG} are also reported. It is clear that tests perform differently depending on the type of attack. The negative kurtosis attack, which was the easiest to detect with the NIST tests, is now the hardest as it alters the distribution the least. In order to get conclusive results with low pass rates for all \texttt{PRNG}s, the sample size had to be set to at least $10^5$. This highlights the importance of using both types of testing: on the random bit stream and after transformation, as different types of attacks may be detectable.